\documentclass[twocolumn]{svjour3}
\usepackage{xcolor,colortbl}
\usepackage{ucs}
\usepackage[utf8x]{inputenc}
\usepackage{graphicx}
\usepackage{epstopdf}
\usepackage{bm}
\usepackage{amsmath}

\usepackage{subfig}
\usepackage{datatool}
\usepackage{filecontents,pgfplots}
\usepackage{pgfplots}
\usepackage{mathtools}
\usepackage{pgf,tikz}

\usepackage{lmodern}
\usepackage{float}
\usepackage{longtable}
\usepackage{pdflscape}
\usepackage{multicol}
\usepackage{graphicx}
\usepackage{epstopdf}
\usepackage{url}

\usepackage{amssymb,amsmath,amstext,amsthm}
\usepackage{enumerate}
\usepackage{listings}
\usepackage{subfig}
\usepackage{datatool}
\usepackage{filecontents,pgfplots}
\usepackage{pgfplots}

\usepackage{algorithm}
\usepackage[noend]{algpseudocode}

\begin{document}
\sloppy
\definecolor{revision}{HTML}{AA0000}
	
	
	\title{Towards hybrid primary intersubjectivity:\\ a neural robotics library for human science}

	\author{Hendry F Chame${}^\dagger$  \and Ahmadreza Ahmadi${}^\dagger$ \and Jun Tani${}^\dagger$}
	
	
	\institute{${}^\dagger$Cognitive Neurorobotics Research Unit (CNRU).Okinawa Institute of Science and Technology Graduate University (OIST). 1919-1 Tancha, Onna-son, Okinawa, Japan 904-0495.\\ \email{\{hendryfchame,ar.ahmadi62,tani1216jp\}@gmail.com}	}
	
	
	\journalname{myjournal}
	\date{}
	\maketitle	
	
	\begin{abstract}
	Human-robot interaction is becoming an interesting area of research in cognitive science, notably, for the study of social cognition. \textit{Interaction theorists} consider \textit{primary intersubjectivity} a non-mentalist, pre-theoretical, non-conceptual sort of processes that ground a certain level of communication and understanding, and provide support to higher-level cognitive skills. We argue this sort of low level cognitive interaction, where control is shared in dyadic encounters, is susceptible of study with neural robots. Hence, in this work we pursue three main objectives. Firstly, from the concept of \textit{active inference} we study primary intersubjectivity as a second person perspective experience characterized by predictive engagement, where perception, cognition, and action are accounted for an hermeneutic circle in dyadic interaction. Secondly, we propose an open-source methodology named \textit{neural robotics library} (NRL) for experimental human-robot interaction, and a demonstration program for interacting in real-time with a virtual Cartesian robot (VCBot). Lastly, through a study case, we discuss some ways human-robot (hybrid) intersubjectivity can contribute to human science research, such as to the fields of developmental psychology, educational technology, and cognitive rehabilitation.
	
	\end{abstract}
	
	\keywords{Social cognition \and Interaction theory \and Neural robotics \and Human-robot interaction \and Predictive engagement \and Free energy principle \and Developmental psychology \and Educational technology \and Cognitive rehabilitation}
		
	\section{Introduction}
\label{intro}

At present, technology has permeated distinct spheres of human society, which has fundamental implications for the research of cognition. Considering the field of robotics, in order to cope with challenges of our times, it is important to study how the inclusion of robots can transform the economic and social organization of our society (Granulo et al. \cite{granulo2019}), and possible ways of dealing with those changes. It is also crucial to explore how human-robot interaction research can serve beneficial purposes, such as helping in advancing the state of the art in several fields of cognitive science. 

In this evolving context, several research communities with distinct purposes have dedicated themselves to study human-robot interaction in recent years. Perhaps myriads of nominal categories could be proposed to organize contributions in the field, some examples are the kind of robot involved (e.g. humanoids, pets, exoskeletons) and the goals of interaction (e.g. educational, assistance, collaboration). In view of such diversity, a convenient point to start from is structuring our discussion around a key element of analysis that can be systematically revisited. Hence, we propose to deal with an important question evoked during interdisciplinary conferences and encounters in our field, which is simply stated as: \textit{who is the subject in the interaction?}

\subsection{The human subject}

Continent upon the quality of subjectivity is placed on the human partner side, synthetic agents have served methodological purposes for varied ends, such as diagnosing, assisting, and learning. Hence, by providing assistance therapy, robots can help humans recover from sensory-motor deficits in some neuro-psychological conditions (Gassert \& Dietz \cite{gassert2018}). Interaction has also served to study neural development in \textit{autistic spectrum disorder} (ASD, Robins et al. \cite{robins2005}, Scassellati et al. \cite{scassellati2012}, Ismail et al. \cite{ismail2019}). Furthermore, robotics has contributed to the acquisition of computational thinking skills (Atmatzidou \&  Demetriadis \cite{atmatzidou2016}), improving executive functions of planning and control (Di Lieto et al. \cite{DiLieto2020}), sensory-motor gaming (Kose-Bagci \cite{kose2009}), and metacognitive and problem solving (Atmatzidou el al. \cite{atmatzidou2018}). 

An important aspect to point out is that diverse approaches can be adopted for providing the robot with control and interaction capabilities. Since the focus is placed on the human subject, the internal structure of the artificial partner is considered of secondary importance to the analysis of experimental protocols, which include mostly observations on the human side. Frequently, the robot is included as a black-box system, providing some degree of standardization to ground conclusions. Although this sort of studies is relevant for advancing several fields in human science, we are interested in considering artificial subjectivity in the research hypotheses design, which is described next.

\subsection{The robot subject}

Focusing on the robot as a subject originates from the interest in studying human cognition. Initiatives under this perspective share the expectation that plausible theories in cognitive science can be constructed through modeling and implementing cognitive control schemes in robots. Here, the human partner acts by stimulating the robot during interaction. Thus, observations on the human side are considered in terms of evaluating experimental hypotheses on the synthetic prototype. 

In order to extend the result of robotics research to human science, some authors have followed, from an interdisciplinary perspective, a behavior-centered approach, so focusing on robotics research results that could be directly mapped to analogous studies and tasks with human infants and children (Cangelosi \& Schlesinger \cite{cangelosi2015}). According to Asada et al. \cite{asada2009}, since the knowledge available in developmental cognitive robotics is insufficient, several implementations are proposed to investigate cognitive functions from the designers' limited understanding of them (e.g. to study foundations of communication, Kuniyoshi et al. \cite{kuniyoshi2004}).

The research conducted in our lab has focused on composable continuous state space neuro-dynamic structures for the study of cognition (Tani \cite{tani2016}). Taking inspiration from brain sciences, several recurrent neural network architectures (e.g. Tani \cite{tani2003} and  Murata et al. \cite{murata2013}) have been proposed for learning spatial-temporal relations in behavior sequences. These models have been used to investigate diverse skills, such as goal-directed planning (Jung et al. \cite{jung2019}) and imitation (Hwang et al. \cite{hwang2018}, Ohata \& Tani \cite{ohata2020}). Within the neural robotics conception adopted in our lab, cognition is studied as an emergent phenomenon that results from the interaction between a top-down process, that characterizes subjective deliberation and agency, and a bottom-up information process, that accounts for the perceptual reality. Next, we argue neural robots are valuable means to study human-robot non-verbal communication, where both partners are included as subjects in the research hypotheses design.

\subsection{Hybrid primary intersubjectivity}

\textit{Intersubjectivity} is a concept that carries deep philosophical roots. It is fundamental to Hursserl’s foundations on phenomenology \cite{husserl2013}, and received important contributions from Heidegger (with the notion of \textit{being-in-the-world} \cite{heidegger1962}), Merleau-Ponty (through the study of perception and embodiment \cite{merleau1996}), and Habermas (in the \textit{theory of communicative action} \cite{habermas1984}), among several other sources. Dealing with the philosophical complexities of the concept would certainly surpass our current scope. Thus, our focus is rather placed on the much more circumscribed sphere of social psychology.

A conceptual distinction in social cognition research has been established to define the sort of access a person uses in understanding another person. According to Fuchs \cite{fuchs2013}, from an experiential level of analysis, the access possibilities to oneself and others conform the triad: first (1PP), second (2PP), and third (3PP) person perspective. Hence, subjective experiences are accessible from 1PP, co-experiences or intersubjective experiences (reciprocal interaction, forms of mutual relatedness) are accessible from 2PP, and one-way, vicarious, or remote observations are accessible from 3PP. An intense debate has been established among theorists concerning the experiential level of analysis from which studying intersubjectivity. This is reviewed next in order to delimit the theoretical scope of this work.

Several authors (e.g. Gallagher \cite{gallagher2001}, Fuchs \cite{fuchs2013}, Newen \cite{newen20184E}) have pointed out that studies in social cognition have traditionally explained how individuals understand and interrelate with each other from the perspective of theory of mind. Notably, under the approaches of \textit{theory theory} (TT) and \textit{simulation theory} (ST). In essence, from a 3PP experiential level of analysis, TT theorists investigate intersubjective relations as specialized cognitive abilities for explaining and predicting behavior, based on the employment of folk psychological theories about how behavior is informed by mental states (e.g. Premack \& Woodruff \cite{premack_woodruff_1978}, Welman \cite{wellman1992}, Gopnik \& Schulz \cite{gopnik2004}, Leslie \cite{Lesley1987}). ST theorists study intersubjectivity from a 1PP experiential level of analysis, as how  mental experience becomes an internal model for understating the other's mind, so thoughts or feelings of the other person are simulated as the subject would be in the other's situation (e.g. Gallese \& Goldman \cite{gallese1998}, Rizzolatti \& Fogassi \cite{rizzolatti2014}, Davies \& Stone \cite{davies1995}, Goldman \cite{goldman2006}).

Developmental psychology research has challenged theory of mind accounts of social cognition, and described primary and secondary intersubjectivity stages of development (Gallagher \cite{gallagher2008a}, Spaulding \cite{spaulding2012}). Thus, \textit{primary intersubjectivity}, originally described by Trevarthen \cite{trevarthen1979}, consists in a voluntary interpersonal communication process characterized by intentionality and adaptation, which is founded in innate, embodied, pre-theoretical, non-conceptual fundamental capacities for self-expression and understanding others (e.g. facial gesticulation; proprioceptive sensation, automatic attunement, detection of intentional behavior, eyes motion tracking, noticing emotions in gestural intonation and expression, among others). \textit{Secondary intersubjectivity} is theorized to be constituted later, transcending the face-to-face sort of interactions to a context of shared attention, mediated by communication about objects and events in the environment. Conforming to Rochat \& Passos-Ferreira \cite{Rochat2009}, the stage of \textit{tertiary intersubjectivity} is characterized by processes of negotiation with others about the values of objects, from shared and self representations. Our work is interested in studying social interaction, as characterised by primary intersubjectivity.

Interdisciplinary efforts from developmental psychology, phenomenology, and philosophy of mind, have constituted, in the last decades, a diverse field of research that investigate four central features of cognition. Thus, according to Newen et al. \cite{newen20184Eb}, \textit{4E cognition} is considered to be \textit{embodied}, \textit{embedded}, \textit{extended}, and \textit{enactive}. Theorist in 4E cognition research cognitive phenomena as dependent on the body characteristics (on its physiology, biology, and morphology), on the particular structure of the environment (e.g. natural, technological, social), and on the active embodied interaction of the agent with the environment.

When considering the study of 4E properties of social cognition, the critical movement against theory of mind approaches was very  much influenced by the proposal of  \textit{interaction theory} (IT, Gallagher \cite{gallagher2001}, also named \textit{embodied social cognition} Gallagher \cite{gallagher2008a}). For IT, experiencing the feelings and intentions of another person is mostly accounted for by a 2PP access to immediate perception of embodied interaction with others, which constitutes a simpler, non-mentalistic, on-line capacity. Hence, mind-reading skills (as studied in TT and ST) consists in specialized forms of intellectual activity less regularly used when basic embodied processes fail to account for a given situation. Although IT is not an uniform theoretical field, according to Newel \cite{newen20184Eb} theorist share the two following ideas: a) understanding others does not involve observing others on the regular basis, but interacting with them, and b) the experiential access in understanding through interaction is immediate or direct perception. We study primary intersubjectivity in the context of IT.

When analysing differences between IT theorists, it is fundamental to observe the peculiarities or sort of interaction that constitute a particular research scope. For enactivist theorists (e.g. De Jaegher \& Di Paolo \cite{Jaegher2007}), social interaction is characterized by coupling, which maintains an identity in the relational domain, and by individual autonomy. An example would be walking in the opposite direction in a narrow corridor, where individuals are attempting to stop interacting but the interaction self-sustains notwithstanding their will. Conforming to Reddy \cite{reddy2010}, it is not the structure of the situation that determines a 2PP level access experience in the interaction, but the fact of mutual acknowledgment, emotional involvement, or engagement (e.g. sharing a smile, attraction, interest, surprise). 

Our research scope is concerned with engaged interactions as described by Reddy. A distinguishable aspect of our work is the interest in direct interaction experiences characterized by intention and purpose. That is, along with possessing means for adaptation or fitness to other's actions, individuals are also capable of employing volitional resources to express their intention. Consequently, dynamic control is shared within the dyad. We believe that this sort of exchange is possible when both agents are capable of, among several skills, processing feedback and formulating proactive expectations on how the situation would look like while enacting in the dyad (Tani \cite{tani2016}). 

In the study of intersubjectivity, IT theorists have disagreed on the importance of knowledge representation and prediction. Conforming to Schlicht \cite{schlicht20184E}, some theorists have radically departed in a non-representational approach and studied emergent interaction from dynamic system theory, whereas moderate theorists have retained mental representation in theorizing social cognition. Examples of the former are studies of minimalistic interaction (Auvray et al. \cite{auvray2009}, Froese \& Ziemke \cite{froese2009}, Lenay \& Stewart \cite{lenay2012}, Froese et al. \cite{froese2014}). Concerning moderate views, according to Gallagher \& Allen \cite{gallagher2018}, some interpretations of the general framework of the predictive model are consistent with methodological individualism (e.g. Hohwy \cite{hohwy2013}), whereas others are consistent with autopoietic enactivist theories (e.g. Kirchhoff \cite{kirchhoff2018} \cite{kirchhoff20184E}). Notably, \textit{free energy principle} (FEP) theory (Friston \cite{friston2013}). 

This work investigates primary intersubjectivity from the perspective of FEP theory, within the scope of 4E social cognition. In agreement with Allen \& Friston \cite{allen2018}, we study FEP theory as a synthetic account to explain the constitutive coupling of the brain to the body and the environment. Thus, internal representation and prediction, in the generative sense, are considered to emerge from the organismic \textit{autopoietic}\footnote{Understood as the persistence of self-organization of an organism given its characteristic dynamical structure, while interacting with the environment.} self-organization. Given the assumption of \textit{ergodic}\footnote{\textit{Ergodicity} implies that the average probability of a system being in a given state is equivalent to the probability of being in such state when observed randomly.} dynamical interchange between the agent and the environment, we investigate perception, cognition, and action as explained by an enactive hermeneutic circle taking place in dyadic encounters (Gallagher \& Allen \cite{gallagher2018}). For this, we study interaction as a process where deliberative control and automatic adjustment coexist. 

A final aspect to discuss, when deconstructing the title semantics, is the meaning of the term \textit{hybrid}. It accounts for the study of social interaction between two partners distinct in nature. In this sense, an important issue investigated in human-robot interaction has been human engagement, where the \textit{uncanny valley effect} (i.e. emotional response in subjects from perceived human resemblance of synthetic objects, Mathur \& Reichling \cite{mathur2016}) has been reported. Differently from this line of research, we focus on the inclusion of robots \textit{as they are} to study interaction, and not on how the robot could substitute the human partner. As elaborated in the next sections, we propose a methodology to model artificial agents for social interaction. Through the presentation of a case study experiment, we discuss some perspectives for the research in human science.

\section{Human/neural-robot interaction}

Neural robotic agents are inspired in brain science research. It is generally considered that understanding brain functions requires the integration of knowledge at multiple levels of abstraction (Hawkins \& Blakeslee \cite{hawkins2007}, Freeman \cite{freeman2014}, Ishii et al. \cite{ishii2011}). Thus, several works have researched synaptic molecular protein synthesis, how neuromechanical signals are transmitted, how spiking activity in a single neuron unfolds, local cell assembly circuits, and the whole brain network. Each of these levels of abstraction are the subject matter of particular disciplines in cognitive science, such as neuroscience, psychology, and artificial intelligence, among others.

We are interested in the study of relational and organizational aspects of cognition by taking a synthetic approach. Particularly, we use recurrent neural networks (RNN), which are highly adaptable nonlinear dynamical systems able to deal with both temporal and spatial information structures. Several architectures have been investigated in our lab (e.g. \textit{continuous time recurrent neural network} CTRNN, Beer \cite{beer1995}, and \textit{multiple timescale recurrent neural network} MTRNN, Yamashita \& Tani \cite{yamashita2008}). More recently, a variational framework inspired by FEP theory named \textit{predictive-coding-inspired recurrent neural network} (PV-RNN, Ahmadi \& Tani \cite{ahmadi2019}) was proposed. This framework is selected as a case study in this work.

Previous research has studied social cognition from FEP theory. Thus, several works (e.g. Van de Cruys et al. \cite{van2014}, Hohwy \& Palmer \cite{hohwy2014}, and Lawson et al. \cite{lawson2014}) have attempted to explain social behavior in ASD through the predictive model account. Commonly, a theory of mind stance has been adopted. This has been also the case for the research of intersubjectivity and communication (e.g. Friston \& Frith \cite{friston2015b}). In general, studying social cognition is tremendously challenging from the methodological point of view. Regularly, researchers have resorted to off-line computer simulations (e.g. Friston \& Frith \cite{friston2015a}), or to on-line indirect interaction mediated by virtual systems, as a means to exert control over extraneous experimental variables (e.g. the technique \textit{hyperscanning}, Babiloni \& Astolfi \cite{babiloni2014}).

A peculiarity that emerges in dyadic direct interaction is that actions depend on both partners' interventions. Moreover, several sensory and motor organs are simultaneously involved and not only voluntary control, but spontaneous or covered reactions are produced and regulated by the central and peripheral nervous systems. Noticing this has been particularly important for the study of neural development in ADS (Torres et al. \cite{torres2013}). Under such complex scenario, we investigate intentional human-robot interaction as a second person perspective access experience.

Human-robot interaction dynamics and the theoretical assumptions adopted in this work are represented in Fig. \ref{fig:interaction}. The neural robot is dotted with the capacities of agency (or deliberation) and compliance (reactiveness, adjustment) in relation to the human partner's actions. In agreement with FEP theory (Friston \cite{friston2011}), the robot's deliberative control is studied as a variational optimization process that involves a hierarchical dynamical representation, in which a top-down information flow (developed by a generative process) characterizes the agency of purposeful actions (or intention to behave) in the interaction context, and a bottom-up information flow (an inference process) accounts for their consequences; so conflicts possibly appearing between these two processes are attempted to be reduced through minimizing free energy as a statistical quantity.

\definecolor{plotRed}{HTML}{8A1227}
\definecolor{plotDarkTeal}{HTML}{204651}

\begin{figure}[th]
	\begin{scriptsize}			
		\begin{tikzpicture}
		\node [] at (0,0){
			\includegraphics[width=0.45\textwidth,keepaspectratio]{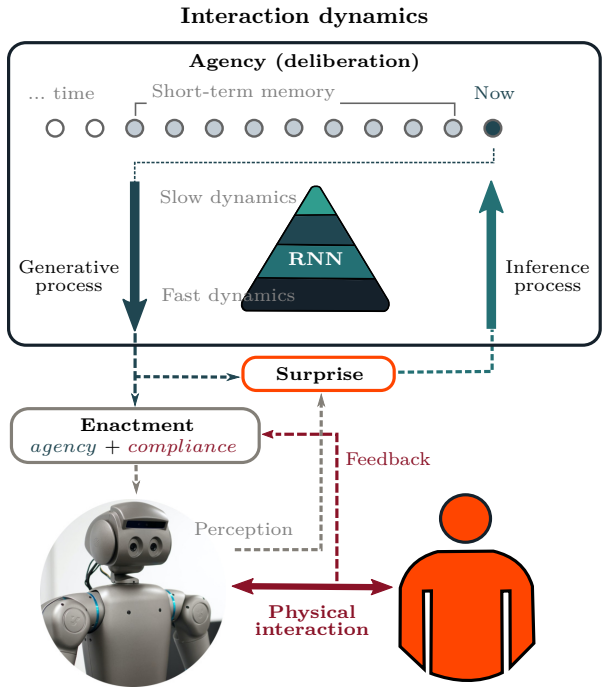}};
		
		\node [color=black] at (0.0,4.6) {\small \textbf{Interaction dynamics}};			
		\node [color=black] at (0.0,4.0) {\scriptsize \textbf{Agency (deliberation)}};
		\begin{scope}[shift={(0.0,-0.7)}]

		
		\node [color=plotDarkTeal] at (2.5,4.3) {\scriptsize Now};
		\node [color=gray] at (-0.8,4.3) {\scriptsize Short-term memory};
		\node [color=gray] at (-3.2,4.3) {\scriptsize ... time};
		\node [color=gray] at (-1.0,2.9) {\scriptsize Slow dynamics};
		\node [color=white] at (0.15,2.1) {\scriptsize \textbf{RNN}};
		\node [color=gray] at (-1.0,1.6) {\scriptsize Fast dynamics};
		\node [color=black] at (-3.1,2.0) {\scriptsize Generative};
		\node [color=black] at (-3.1,1.7) {\scriptsize process};
		\node [color=black] at (3.2,2.0) {\scriptsize Inference};
		\node [color=black] at (3.2,1.7) {\scriptsize process};
		\node [color=plotRed] at (1.1,-0.55) {\scriptsize Feedback};
		\node [color=black] at (0.2,0.55) {\scriptsize \textbf{Surprise}};
		\node [color=gray] at (-0.8,-1.5) {\scriptsize Perception};
		\node [color=black] at (-2.2,-0.1) {\scriptsize \textbf{Enactment}};
		\node [color=black] at (-2.25,-0.4) {\scriptsize \textcolor{plotDarkTeal}{$agency$} + \textcolor{plotRed}{$compliance$}};
		\node [color=plotRed] at (0.1,-2.6) {\scriptsize \textbf{Physical}};
		\node [color=plotRed] at (0.1,-2.8) {\scriptsize \textbf{interaction}};
		\end{scope}
		\end{tikzpicture}
	\end{scriptsize}
	\caption{Control is shared in the interaction based on the individuals' capacities of enacting agency and compliance. The capacity of agency is studied through the optimization of free energy (\textit{surprise}) in short-term memory of direct perception. Hence, from the robot's unique past, an intended future is reinterpreted in the hierarchical RNN for deliberation control and projected in the perceptual space. The capacity of compliance is modulated as a proportional integral (PI) control scheme for automatic adjustment of the body to the actions induced by the partner.}
	\label{fig:interaction}
\end{figure}

Compliance is a fundamental capacity for social interaction which is based on the awareness of others intentions and knowing what those intentions refer to. This capacity has been studied from a developmental perspective at its early emergence (Reddy \cite{reddy20184E}, Reddy et al. \cite{reddy2013}). When contextualizing our work in the IT literature, we start from the assumption that automatic processes are available to the person, so the body reacts by adjusting to some extend to the partner's actions. Here, we do not study nor evaluate this assumption, this is an aspect that remains for a further examination. Thereby, we focus on understanding deliberation in the dyad from our interpretation of FEP principle theory, taking automatic adjustment as granted. 

Our previous research (Chame \& Tani \cite{chame2019}) has shown in on-line human-robot interaction a possible way in which compliance, as a cognitive (volitional) dynamic process, can relate to reactive motor adjustment. Hence, \textit{cognitive compliance} was studied as a dynamical integrative optimization process that characterizes how deliberation is influenced by sensory stimulation induced by the partner's actions in the dyad. Thus, it was shown how an agent with strong belief tends to act egocentrically to the environment without changing its internal state or intention, whereas an agent with weak belief tends to act adaptively to the environment by easily changing its internal state or intention, while the capacity of adjustment was kept constant.  

We consider that the perspective adopted here for studying primary intersubjectivity is consistent with the characterization provided by Trevarthen \cite{trevarthen1979}, who views this construct as a deliberative-compliant process. Therefore, we propose to investigate enactivist social cognition in human-robot interaction as a relevant methodological approach for several fields of human science research, which is discussed in Sec. \ref{sec:perspectives}. Next, within the context of interdisciplinarity, fundamental concepts in our field are introduced with minimal mathematical formulation, under the modality of a questions \& answers panel section. 

\subsection{What is free energy?}

Free energy principle theory was initially proposed as a unifying framework in brain science (Friston \cite{friston2010}). More recently, the theory has been extended and related to the fields of theoretical biology, statistical thermodynamics, and information theory (Friston. \cite{friston2013}). A core assumption in FEP theory is that living organisms are driven by the tendency to resist the second law of thermodynamics, so to maintain their internal structure or dynamics in a constantly changing environment. 

Conforming to Allen \& Friston \cite{allen2018}, from this fundamental drive for existence, biological system are characterized by the following properties: \textit{ergodicity} (an organism occupies or revisits some characteristic states more than others over time in order to live), \textit{Markov blanket} (a mathematical description of the boundaries between the organism and the environment, such description is undertaken at multiple levels of analysis), \textit{active inference} (perception and action are locked in a circular causality relation), and \textit{autopoiesis} (emergent and self-organized maintenance of organismic dynamic structure while interacting with the environment).

FEP theory consists in a predictive account of the mind. It considers that an agent proactively anticipates sensation from empirical priors (in a generative sense), and minimizes free energy as a measure (an upper bound) of \textit{surprise}. As a consequence, the internal state of the agent is maintained within characteristic or habituated regions. Thus, less/more sensory surprise means lower/higher free energy, with more/less likelihood of internal dynamics unfolding in a given region.

\subsection{How is free energy minimized?}

Generally, free energy is considered to be minimized in two fundamental ways: through \textit{predictive coding}, for vicarious perception, and through \textit{active inference}, for goal directed action (Friston et al. \cite{friston2011}). In predictive coding, surprise is minimized in a bottom-up pathway, so the internal state is modified to generate more consistent predictions with respect to sensory evidence. This principle has been explored in theory of mind studies of social cognition (e.g. Van de Cruys et al. \cite{van2014}, Kilner et al. \cite{kilner2007}, Koster-Hale \& Saxe \cite{koster2013}). Differently, in active inference sensory predictions are attempted to be fulfilled in a top-down pathway, by taking purposeful actions in the environment (Friston et al. \cite{friston2010a}). 

Regardless of the way surprise is minimized, in FEP theory it is assumed that sensation prediction is accompanied with an estimate of precision. Hence, surprise is considered to be minimized in relation to (divided by) precision, which means that free energy is minimized more when associated with a high precision estimate (i.e. a strong belief). A practical implication of the previous statement is that an agent with strong belief tends to act egocentrically to the environment without changing its internal state or intention, whereas an agent with weak belief tends to act adaptively to the environment, by easily changing its internal state or intention. 

This work focuses on the study of enactivist social cognition based on active inference. As pointed out by Friston et al. \cite{friston2011}, in active inference no distinction is established in terms of sensory or motor representations, since motor control signals are considered to be directly generated by proprioceptive predictions, so the individual perceives relevant action affordances in the interaction context. This is going to be discussed in more details when considering the matter of perception as a direct experience. In the appendix sections, active inference is described within the scope of our case study, which includes the PV-RNN framework. The minimization of free energy is equivalent to the maximization of the sensory evidence lower bound (ELBO), which is discussed in the next question.

\subsection{What is the evidence lower bound?}

The ELBO is a quantity introduced in the variational Bayesian (VB) optimization literature. In FEP theory, this quantity corresponds to the summation of two terms, namely the \textit{accuracy} term, representing the prediction error, or surprise; and the \textit{complexity} term, representing the complexity of the internal representation. Mathematically, accuracy is the expected logarithm likelihood of the model with respect to the approximated posterior distributions, and complexity is the Kullback-Leibler divergence (KL divergence) between the approximated posterior and the prior hidden distributions. Therefore,  by definition $ELBO \leq 0$ is an upper limit for the logarithm of the marginal likelihood in the anticipation of the sensory state by the generative process  (see the mathematical details in the Appendix section). For human-robot interaction experiments it is perhaps more intuitive to relate surprise to the negative evidence lower bound (N-ELBO), since increases in N-ELBO correlate with situations where mismatch increases between the anticipation and actual sensation. 

\subsection{Is perception a direct experience ?}

In the study of perception-action, two schools have contrasted, namely, the \textit{contructivist} and the \textit{ecological} theory of perception. These theories have developed over several years. Here, only their basic characterization is presented. Thus, constructivist theorists are very much influenced by Helmholtz's notion of \textit{unconscious conclusion} (Helmholtz \cite{Helmholtz2013}). It is generally assumed that stimulation reaching the sensory apparatus is not sufficient for perception, so intermediate processes (e.g. memory, perceptive schemes, previous experience) intervene between sensation and perception, which characterizes an inferential and indirect process. Contrarily, the ecological school, under the influence of Gibson's studies of visual perception (Gibson \cite{gibson2014}), consider that information available in the ambient suffices, since what is perceived are changes over time and space (an information flow). Accordingly, individuals perceive \textit{affordances} (i.e. functional utilities with respect to themselves and their action capabilities) of objects or the environment. Consequently, perception is studied under the ecological school as a direct or immediate process.

Constructivist theories of perception have been associated with cognitivism, when attributing a predominant role to the brain in the top-down information processing of sensory data (e.g. Gregory \cite{gregory1974}), which largely neglects the richness of information available at the sensory level. However, although neurophysiological, neuropsychological, and psychophysical scientific evidence has supported the coexistence of constructivist and ecological perceptive processes in the brain (Norman \cite{norman2002}), according to De Wit \& Withagen \cite{deWit2019}, ecological psychologists have been criticized for ignoring the brain in their theoretical formulations. 

More recently, Linson et al. \cite{linson2018} have argued that when replacing traditional inferential explanations, based on the notion of passive input, with the notion of active input, ecological theorists still maintain an input-output account of perception. According to the authors, active inference is characterized in FEP theory as direct engagement (in the thermodynamic sense) between the agent's sensory system and the environment. Thus, sensation is considered to be anticipated from a generative process, which does not constitute an inferential process in the input-output sense. As pointed out by Bruineberg \& Rietveld \cite{bruineberg2019}, perception and action from anticipatory dynamics and free-energy minimization ensure the agent with the capability of maintaining adaptive sensibility to relevant affordances in a given context. This would not be about reconstructing the structural hidden causes of the environment.

This debate has been certainly inspiring for artificial intelligence and robotics research. It could be relevant bringing up to discussion influential works by Brooks in behavioral robotics (e.g. Brooks \cite{brooks1990}\cite{brooks1991}). The field was also influenced by Braitenberg \cite{braitenberg1986}. In behavior robotic agents, sensory input has been directly mapped to motor output. This idea would arguably conform to principles of the ecological theory of perception (Tani \cite{tani2016}). Although interesting behavior can result from minimal control schemes, this sort of robots are only capable of reacting to the environment. Lacking of intention, they behave in stereotyped ways and exhibit low capacity of generalization, even to slightly different situations.

In the approach adopted in our research, robots are capable of active inference, so they perceive action affordances directly as a vector flow in the perceptual space (i.e. no distinction is established in terms of sensory or motor representations). Hence, such neural robots try to fulfill their intention in the interaction by directly performing afforded actions making sense in a given context. Next, we propose a methodological resource for designing these sort of neural agents for conducting human-robot interaction experiments. 

\section{An open-source library }
\label{sec:nrl}

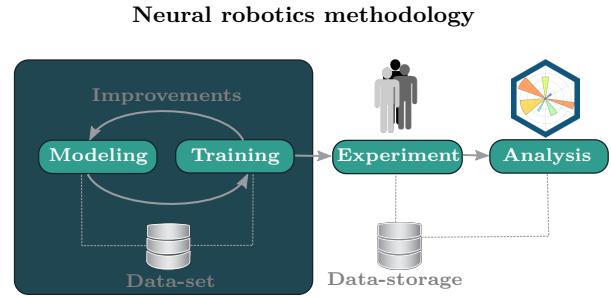
\begin{figure}[th]
	\centering
	\begin{scriptsize}			
		\begin{tikzpicture}
		\node [] at (0,0){
			\includegraphics[width=0.45\textwidth,keepaspectratio]{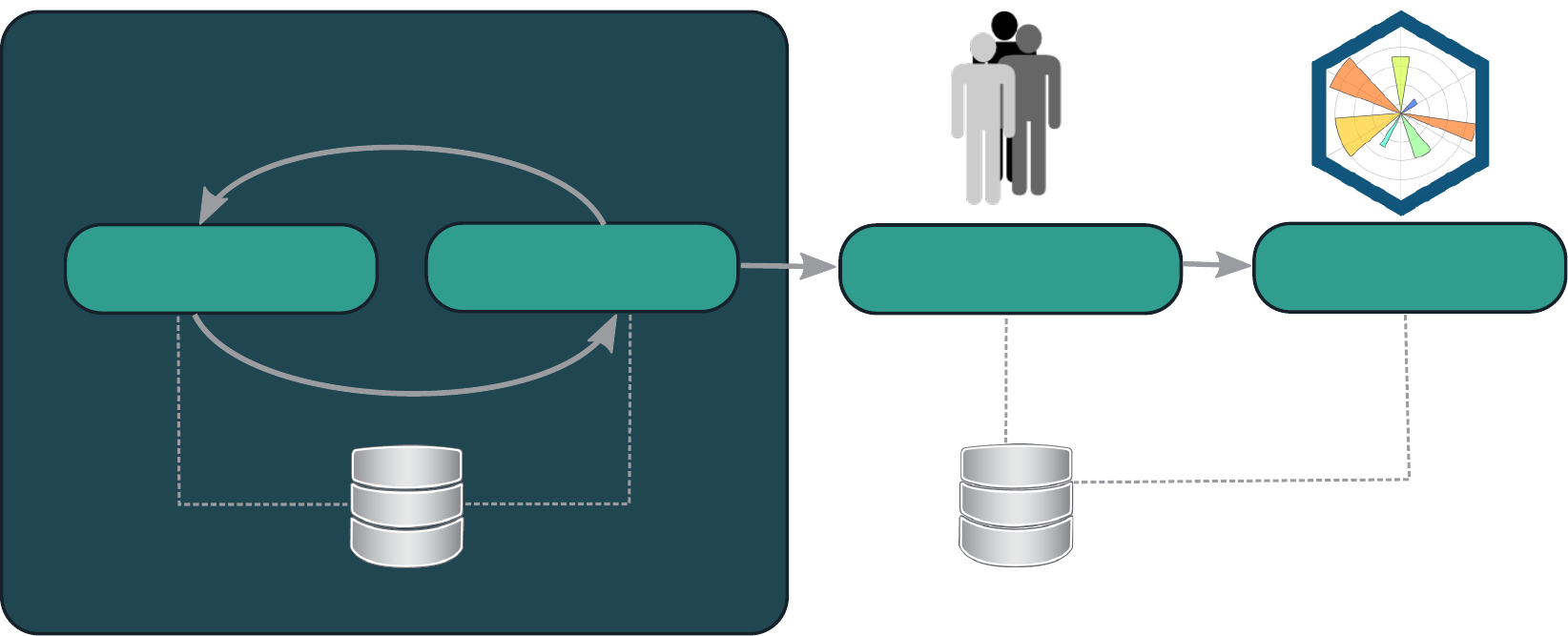}};
		
		\node [color=black] at (-0.1,2.1) {\small \textbf{Neural robotics methodology}};
		\node [color=gray] at (-1.9,1.05) {\scriptsize \textbf{Improvements}};
		\node [color=white] at (-2.8,0.25) {\scriptsize \textbf{Modeling}};
		\node [color=white] at (-1.0,0.25) {\scriptsize \textbf{Training}};
		\node [color=white] at (1.15,0.25) {\scriptsize \textbf{Experiment}};
		\node [color=white] at (3.1,0.25) {\scriptsize \textbf{Analysis}};
		\node [color=gray] at (-1.85,-1.4) {\scriptsize \textbf{Data-set}};
		\node [color=gray] at (1.1,-1.4) {\scriptsize \textbf{Data-storage}};
		\end{tikzpicture}
	\end{scriptsize}
	\caption{The researcher develops a prototype of the neural agent by setting the model's parameters, collecting the behavior dataset, and training the model. In the experimental phase, data from interaction with the subjects is registered, and lately analyzed with the help of graphical and statistical tools.}
	\label{fig:methodology}
\end{figure}

The \textit{neural robotics library} (NRL) is a project designed to serve as an open-source tool for interdisciplinary research in human social cognition. NRL is structured according to the four-step methodology proposed in Fig. \ref{fig:methodology}. The project focuses on providing neural robots with the capability of deliberation control for interaction (see Fig. \ref{fig:interaction}). The software is released at the github repository\footnote{Project repository: https://github.com/oist-cnru/NRL} under the terms of the 3-Clause BSD License. NRL considered recommendations for systems and software engineering developments by the standard ISO/IEC 25010 \cite{iso2011iec25010}.

Prototyping robotic experiments is a challenge endeavor. Provided that robots actuate on the environment, it is fundamental to consider safety, for this, run-time errors must be carefully anticipated and handled. Human-robot interaction presents the supplementary difficulty of requiring computation efficiency for real-time performance. By considering previous experiences with diverse robot tasks (e.g. manipulation in Chame \& Martinet  \cite{chame2015}, human inspired locomotion and object approach in Chame \& Chevallereau \cite{chame2016grounding}, walk with top-down and bottom up visual attention in Chame et al. \cite{chame2016top}), the C++ programming language (Stroustrup \cite{stroustrup2000}) was selected for developing the NRL project. 

In terms of abstraction from the hardware platform resources, C++ is considered an intermediate level language, which is adequate for situations where the programmer intends to control how low level machine resources are used, such as how memory is allocated and information is retrieved. Conveniently, the language syntax conforms to the Object Oriented software engineering paradigm (Meyer \cite{meyer1997}), which favors the quality of software maintainability. 

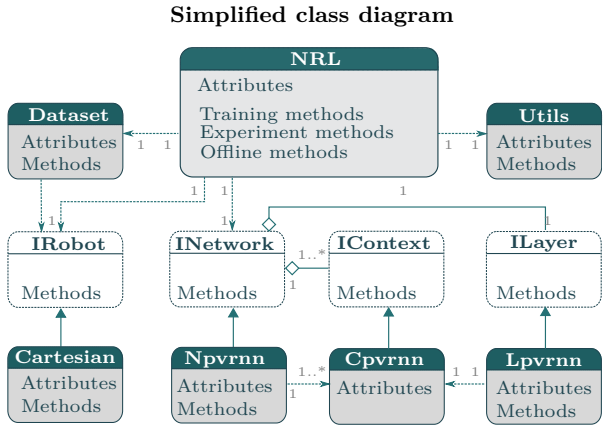
\begin{figure}[th]
	\centering
	\begin{scriptsize}			
		\begin{tikzpicture}
		\node [] at (0,0){
			\includegraphics[width=0.45\textwidth,keepaspectratio]{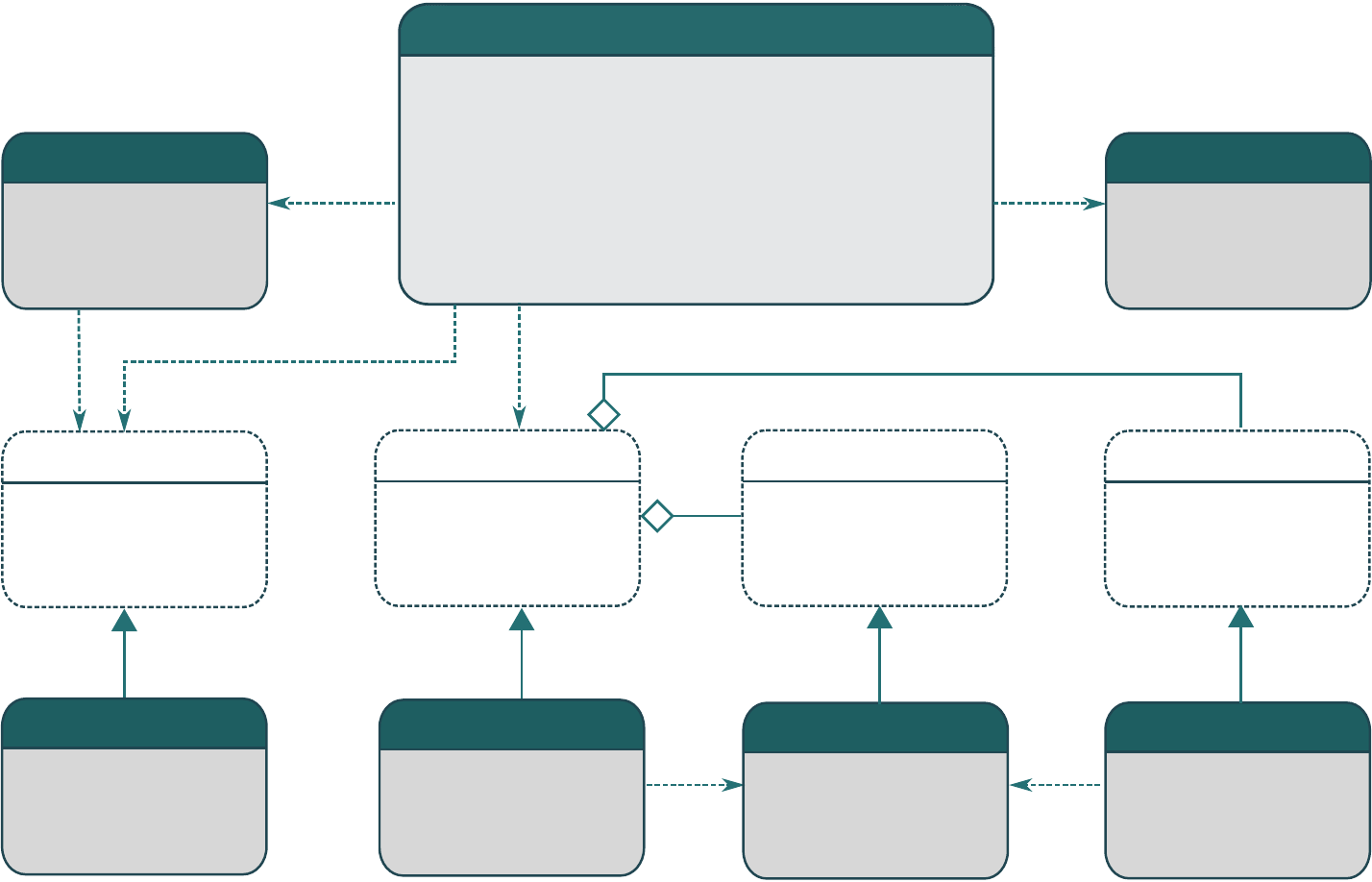}};
		
		\node [color=black] at (0.1,2.9) {\small \textbf{Simplified class diagram}};										
		
		\node [color=gray] at (-1.45,0.6) {\tiny 1};
		\node [color=gray] at (-3.32,0.2) {\tiny 1};
		
		\node [color=gray] at (-1.05,0.6) {\tiny 1};
		\node [color=gray] at (-1.05,0.2) {\tiny 1};
		
		\node [color=gray] at (1.3,0.6) {\tiny 1};
		\node [color=gray] at (3.2,0.2) {\tiny 1};
		
		\node [color=gray] at (1.9,1.2) {\tiny 1};
		\node [color=gray] at (2.25,1.2) {\tiny 1};
		
		\node [color=gray] at (0.1,-0.25) {\tiny 1..*};
		\node [color=gray] at (-0.15,-0.65) {\tiny 1};
		
		\begin{scope}[shift={(0.0,-1.5)}]				
		\node [color=gray] at (0.1,-0.3) {\tiny 1..*};
		\node [color=gray] at (-0.15,-0.6) {\tiny 1};
		\end{scope}
		
		\node [color=gray] at (2.0,-1.8) {\tiny 1};
		\node [color=gray] at (2.25,-1.8) {\tiny 1};
		
		\node [color=gray] at (-1.8,1.2) {\tiny 1};
		\node [color=gray] at (-2.15,1.2) {\tiny 1};
		
		\node [color=white] at (0.15,2.35) {\scriptsize \textbf{NRL}};
		\node [color=plotDarkTeal] at (-0.78,2.0) {\scriptsize Attributes};
		\node [color=plotDarkTeal] at (-0.3,1.6) {\scriptsize Training methods};
		\node [color=plotDarkTeal] at (-0.1,1.35) {\scriptsize Experiment methods};
		\node [color=plotDarkTeal] at (-0.4,1.12) {\scriptsize Offline methods};

		\begin{scope}[shift={(0.0,-0.05)}]
		\node [color=white] at (3.15,1.65) {\scriptsize \textbf{Utils}};
		\node [color=plotDarkTeal] at (3.15,1.3) {\scriptsize Attributes};
		\node [color=plotDarkTeal] at (3.05,1.0) {\scriptsize Methods};
		\end{scope}
		\begin{scope}[shift={(-6.25,-1.75)}]
		\node [color=plotDarkTeal] at (3.15,1.65) {\scriptsize \textbf{IRobot}};
		\node [color=plotDarkTeal] at (3.05,1.0) {\scriptsize Methods};
		\end{scope}
		\begin{scope}[shift={(-4.2,-1.75)}]
		\node [color=plotDarkTeal] at (3.15,1.65) {\scriptsize \textbf{INetwork}};
		\node [color=plotDarkTeal] at (3.05,1.0) {\scriptsize Methods};
		\end{scope}
		\begin{scope}[shift={(-2.1,-1.75)}]
		\node [color=plotDarkTeal] at (3.15,1.65) {\scriptsize \textbf{IContext}};
		\node [color=plotDarkTeal] at (3.05,1.0) {\scriptsize Methods};
		\end{scope}
		\begin{scope}[shift={(-0.0,-1.75)}]
		\node [color=plotDarkTeal] at (3.15,1.65) {\scriptsize \textbf{ILayer}};
		\node [color=plotDarkTeal] at (3.05,1.0) {\scriptsize Methods};
		\end{scope}
		\begin{scope}[shift={(-6.25,-0.05)}]
		\node [color=white] at (3.15,1.65) {\scriptsize \textbf{Dataset}};
		\node [color=plotDarkTeal] at (3.15,1.3) {\scriptsize Attributes};
		\node [color=plotDarkTeal] at (3.05,1.0) {\scriptsize Methods};
		\end{scope}
		
		\begin{scope}[shift={(-6.25,-3.25)}]
		\node [color=white] at (3.1,1.65) {\scriptsize \textbf{Cartesian}};
		\node [color=plotDarkTeal] at (3.15,1.3) {\scriptsize Attributes};
		\node [color=plotDarkTeal] at (3.05,1.0) {\scriptsize Methods};
		\end{scope}
		\begin{scope}[shift={(-4.2,-3.29)}]
		\node [color=white] at (3.15,1.65) {\scriptsize \textbf{Npvrnn}};
		\node [color=plotDarkTeal] at (3.15,1.3) {\scriptsize Attributes};
		\node [color=plotDarkTeal] at (3.05,1.0) {\scriptsize Methods};
		\end{scope}
		\begin{scope}[shift={(-2.1,-3.31)}]
		\node [color=white] at (3.15,1.65) {\scriptsize \textbf{Cpvrnn}};
		\node [color=plotDarkTeal] at (3.15,1.3) {\scriptsize Attributes};
		\end{scope}
		
		\begin{scope}[shift={(-0.0,-3.32)}]
		\node [color=white] at (3.15,1.65) {\scriptsize \textbf{Lpvrnn}};
		\node [color=plotDarkTeal] at (3.15,1.3) {\scriptsize Attributes};
		\node [color=plotDarkTeal] at (3.05,1.0) {\scriptsize Methods};
		\end{scope}
		
		\end{tikzpicture}
	\end{scriptsize}
	\caption{Abstract classes are shown in white background.}
	\label{fig:nrl}
\end{figure}

Figure \ref{fig:nrl} presents the Object Oriented software design. As noticed, in NRL network and layer behaviors are abstracted, respectively, in the INetwork and the ILayer interfaces. In this manner, new functionalities can be added without interfering significantly with available implementations. The network and layers exchange information through the IContext interface. Extended classes would act as data buffers, so information can be shared among other levels of the layered hierarchy. An extended description of the software design is provided in the project's repository. 

In order to ensure compatibility, portability, and reducing the chances of run-time errors, NRL relies on free software platforms developed and maintained independently by third parties. Such resources are largely used in the scientific domain. Hence, since the mathematical modeling of artificial neural networks involves a fair amount of linear algebra, the project considered the Eigen template headers for linear algebra version 3.3 developed at INRIA by Jacob \& Guennebaud \cite{guennebaud2014}. Also, NRL includes the C++ Standard Library, which are available under multiple platforms. 

NRL is a general framework that can be extended to include diverse recurrent neural network architectures. In this work the network type PV-RNN is taken as a case study. The reader interested in consulting the mathematical formalism of PV-RNN is referred to Ahmadi \& Tani \cite{ahmadi2019}. For self-containment, the Appendix sections summarize the mathematical model. In the next section, it is proposed a front-end program that implements a virtual robot, for illustrating NRL operation in on-line interaction. The library has been used in real experiments, as reported in the study by Chame \& Tani \cite{chame2019}, which included the humanoid Torobo.

\section{Interacting with VCBot}
\label{sec:vcbot}

\begin{figure}[h!]
	\centering
	\begin{scriptsize}			
		\begin{tikzpicture}
		\begin{scope}[{scale=0.7}]
		\node [] at (0,0){
			\includegraphics[width=0.45\textwidth,keepaspectratio]{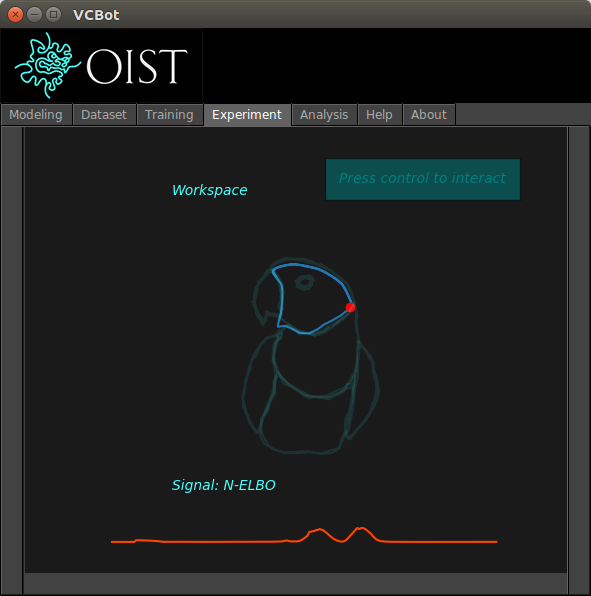}};
		\node [color=black] at (0.0,6.1) {\small \textbf{The graphical user interface}};
		
		\end{scope}
		\end{tikzpicture}
	\end{scriptsize}
	\caption{VCBot in experiment mode ready for interaction. The training dataset (in watermark color for reference) includes behavior primitives globally shaping a macaw cub. The end-effector is represented by the red circle. Recent behavior is shown in light blue (the robot is reproducing the shape of the head). }
	\label{fig:VCBot}
\end{figure}

The virtual Cartesian robot, shortly named VCBot, is a program designed for interaction with an artificial neural agent through the computer mouse. The program is released at the github repository\footnote{Project repository: https://github.com/oist-cnru/VCBot} under the terms of the 3-Clause BSD License. VCBot is modeled as a two degrees of freedom robot articulated by prismatic joints (Khalil \& Dombre \cite{Khalil02}). The software is implemented as a client application in the python programming language (Van Rossum \cite{van2011}), which runs on the top of the NRL back-end. The project is documented in details, with descriptions on the operation modes available. Tutorial videos for installation and operation are proposed in the project's repository. The graphical user interface (GUI) is designed following a notebook layout (see Fig. \ref{fig:VCBot}). It includes in dedicated tabs the four methodological steps shown in Fig. \ref{fig:methodology}, which is discussed below. 

\subsection{Modeling}

Agent modeling starts by the selection of the network features. The PV-RNN parameters shape important qualities such as compliance in the deliberation style of the agent. Unfortunately, the selection of parameters cannot be done analytically, so the method of trial-and-error should be employed. Table \ref{tab:paramsModel} provides a qualitative description of the role of the parameters and some hints on their selection. An important aspect to consider is that computational complexity is conditioned to the number of intermediate layers and the amount of neurons within layers. Therefore, it is recommended to start by profiling a reduced structure and gradually increasing its complexity until it is able to handle the task. This iterative process is represented in Fig. \ref{fig:methodology} as the cyclic flow labeled \textit{improvements}. For reference, the models included for demonstration in VCBot comprise two layers: the Low layer is composed of 40 d units, 4 z units, and the time constant set to 2; whereas the High layer is composed of 10 d units, 1 z unit, and the time constant set to 10. 

\begin{table}[h!]
	\caption{\small{\textbf{PV-RNN parameters selection}}}
	\label{tab:paramsModel}
	\begin{center}
		\begin{tabular}{p{1.4cm} p{6.15cm}}
			\small{\textit{Parameter}}& \small{\textit{Description}} \\ \hline
			\small{z units} & \small{Represent the stochastic latent state in the prior and posterior distributions (see Eqs. \eqref{eq:prior} and \eqref{eq:posterior}). In PV-RNN these units encode a Gaussian distribution parameterized by a mean (or expectation) $\mu$ and a standard deviation $\sigma$, so $z = \mu+\sigma\ast\epsilon$ with $\bm{\epsilon}$ sampled from $\mathcal{N}(0,1)$.}\\\hline
			\small{d units} & \small{Represent the deterministic latent state (see Eq. \eqref{eq:d}). As a rule of thumb they are set ten times more numerous than z units.}\\\hline
			\small{Regulation} $w$& \small{Is a meta-parameter which influences the learning of the posterior and the prior distributions (see Eq. \eqref{eq:elbo}). In general, the higher the parameter is set, the more similar the hidden prior and posterior distributions would be, so the internal representation would be less sensitive to stochasticity during interaction (when deliberating, the agent would comply less to the partner's intentions). On the other hand, if $w$ is set too low, the agent's generative process (based on the prior distribution) would be poor, so deliberation will tend to be erratic}.\\\hline
			\small{Timescale\ \ \ constant} & \small{The timescale conditions the temporal dynamics of the layers. The constant should be selected increasing proportionally between adjacent layers from the lowest to the highest, so low layers present faster dynamics than higher layers. For example, assuming a configuration of three layers, in case $\iota^\mathrm{middle} = 5\iota^\mathrm{low}$ (see Eq. \eqref{eq:d}), then $\iota^\mathrm{high} = 5\iota^\mathrm{middle}$.}\\\hline
		\end{tabular}
	\end{center}
\end{table}

\subsection{Training}

The objective of constituting training sets is to capture fundamental behavior on the robot side for studying during interaction. Thus, depending on the theoretical aspect under consideration, such basic skills could be assumed to be either innate to the agent, or acquired through developmental sensory-motor processes (e.g. through motor babbling, imitation). 

A convenient method used for registering behavior primitives is kinesthetic demonstration, where the human directly moves the robot body to show the behaviors. For the case study, a two-dimension motion primitive set was constituted, inspired by the body shape of a macaw cub (perhaps on the species \textit{ara ararauna}). A total number of seven primitives were included, registered at 100 millisecond sampling period, during 72 time steps, conforming limit cycles in a clockwise sense. These trajectories represented distinct anatomic regions of the bird (i.e. the left eye, the head, the beak, the neck, the right wing, the belly, and the left wing). Only individual primitives were included in the dataset, so the robot did not learn how to relate one primitive to another. Hence, an important aspect to observe is whether possible relations between previous knowledge emerge during interaction. 

\begin{figure}[h!]
	\centering
	\begin{scriptsize}			
		\begin{tikzpicture}
		\node [] at (0,0){
			\includegraphics[width=0.49\textwidth,keepaspectratio]{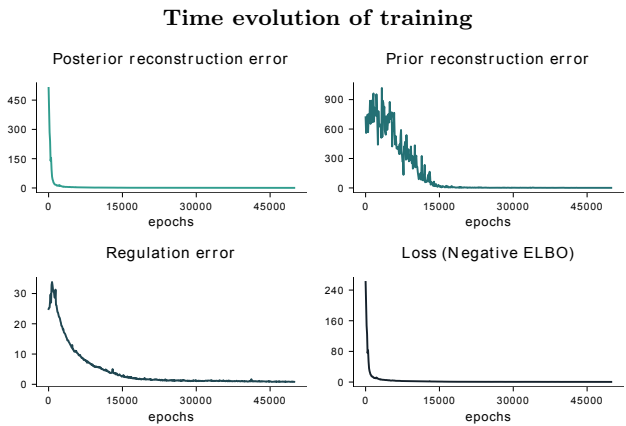}};
		\node [color=black] at (0.0,2.9) {\small \textbf{Time evolution of training}};
		\end{tikzpicture}
	\end{scriptsize}			
	\caption{Training was optimized during 50,000 epochs. The parameter selected for the Adam method were $\alpha=0.001$, $\beta_1=0.9$, $\beta_2=0.999$. The posterior and prior reconstruction errors are calculated by taking the softmax transformation on training data along every degree of freedom of the robot, and computing the sum of the Kullbach-Leiber divergence between the training reference and the output generated by the network from the posterior and the prior distribution respectively.}		
	\label{fig:training}
\end{figure}

The model was trained during 50,000 epochs (see Fig. \ref{fig:training}), following the Adam method for stochastic optimization (Kingma \& Ba \cite{kingma2014}). As noticed on the top row, the network optimized faster the reconstruction error (the accuracy component of the ELBO) related to the posterior distribution, and gradually improved the results for the prior distribution, which can be noticed in the way the signal corresponding to the regulation error (at the bottom-left) decreases. This means a gradual increase of complexity in the model, since the prior and the posterior distributions are becoming more similar. Thus, as seen in the plot on the bottom-right, the negative ELBO is minimized over the training epochs. 

After training, the primitives were generated from the prior distribution to evaluate the quality of behavior achieved. As seen at the top of Fig. \ref{fig:primGeneration}, the generation process could reproduce in overall the body shape of the macaw cub, as compared with the training set trajectory shown in watermark color in Fig. \ref{fig:VCBot}. For observing the internal representation self-organized by the network, two principal component analysis (PCA) from the activity of d units at the High layer (slowest dynamics) were plotted at the bottom of the figure. An important aspect to be noticed is the connectivity between regions in the internal representation. Hence, some primitives are represented inter-connected whereas others are not. The absence of connectivity should imply higher difficulty in switching between those attractor regions, this is going to be discussed in more details when analyzing the interaction results. It is also important noticing that although there may be overlap in the regions, the flow of activity could be in the opposite sense. Since all the primitives were captured clockwise, their internal representation also preserved those spatial-temporal relations. 

\definecolor{primA}{HTML}{67EBEC}
\definecolor{primB}{HTML}{52C6D8}
\definecolor{primC}{HTML}{3EA2C3}
\definecolor{primD}{HTML}{2D7DAF}
\definecolor{primE}{HTML}{1E5C9B}
\definecolor{primF}{HTML}{123E86}
\definecolor{primG}{HTML}{082378}

\begin{figure}[h!]
	\centering
	\begin{scriptsize}			
		\begin{tikzpicture}
		
		\node [] at (1,0){
			\includegraphics[width=0.45\textwidth,keepaspectratio]{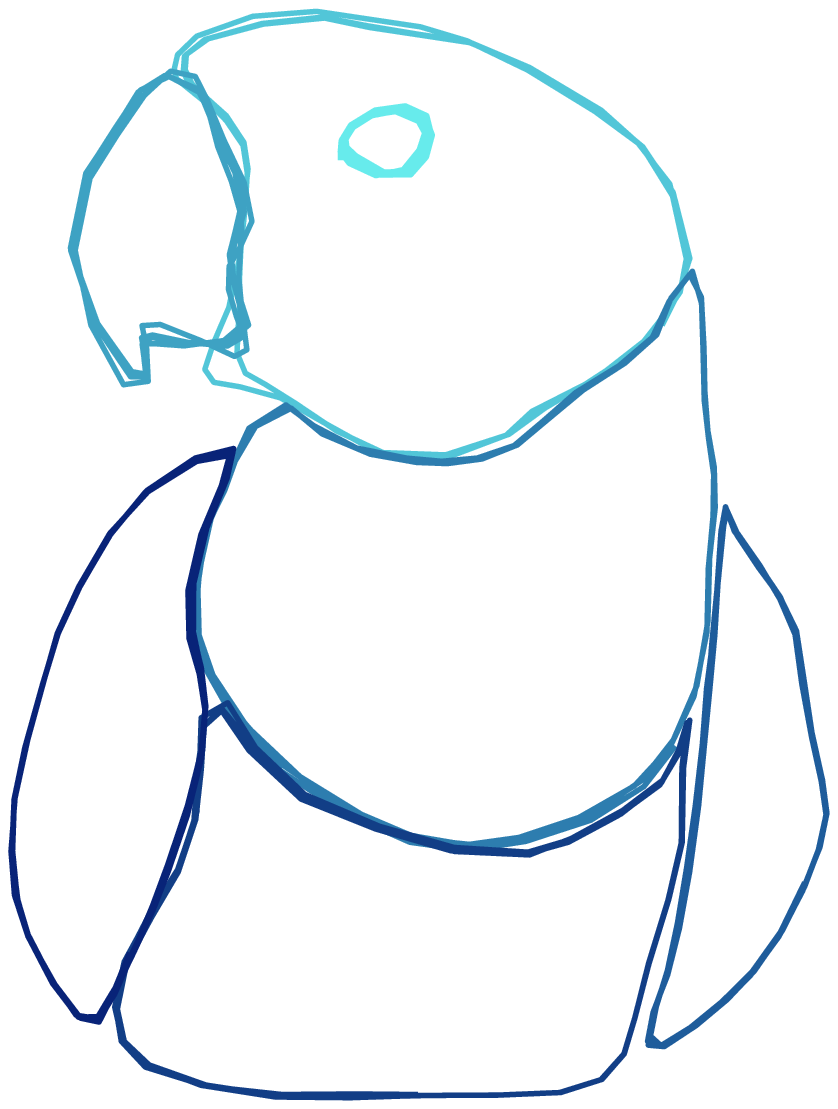}};
		
		\node [] at (1,-6){
			\includegraphics[width=0.45\textwidth,keepaspectratio]{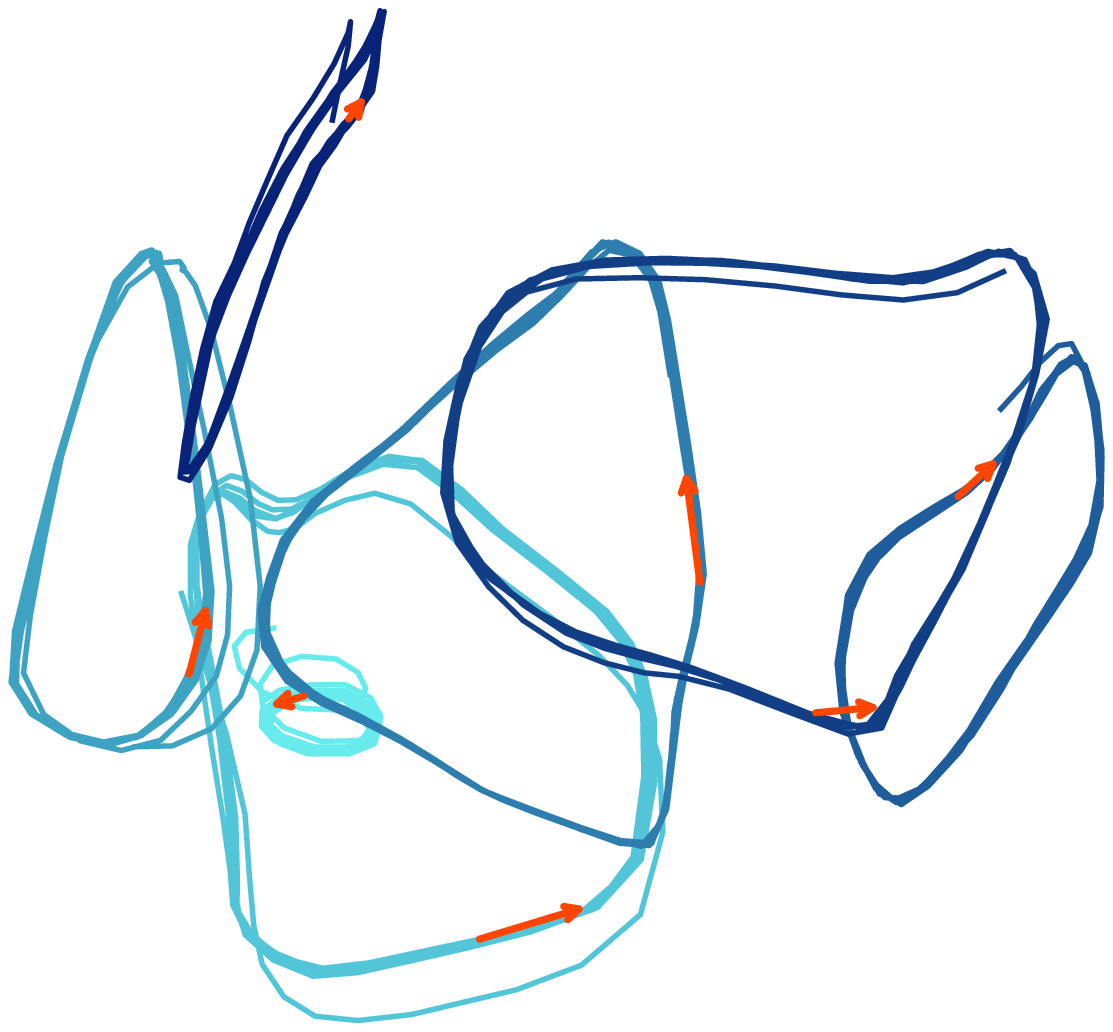}};
		
		\node [color=black] at (1.0,3.2) {\small \textbf{Motion primitives in the workspace}};
		
		\node [color=black] at (1.0,-3.5) {\small \textbf{Two-PCA d states (High Layer)}};
		
		\node [color=primA] at (-0.2,-7.6) {\scriptsize \textbf{Eye}};
		\node [color=primB] at (0.2,-8.2) {\scriptsize \textbf{Head}};
		\node [color=primC] at (-1.2,-7.5) {\scriptsize \textbf{Beak}};
		\node [color=primD] at (0.8,-7.2) {\scriptsize \textbf{Neck}};
		\node [color=primE] at (2.0,-5.4) {\scriptsize \textbf{Belly}};
		\node [color=primF] at (3.0,-7.8) {\scriptsize \textbf{Right-wing}};
		\node [color=primG] at (1.0,-4.0) {\scriptsize \textbf{Left-wing}};
		\end{tikzpicture}
	\end{scriptsize}
	\caption{On the top, the seven behavior primitives representing globally the body shape of a macaw cub were successfully learned by the prior generation process. On the bottom, the resulting internal representation self-organized by the network form training. It is important noticing the connectivity between regions, and the sense of information flow in the attractors developed, which is indicated by the orange arrows.}
	\label{fig:primGeneration}
\end{figure}

\subsection{Experiment}

The goal of the human-robot interaction experiment was to observe mutual interaction and influence between the partners. For this, the human was instructed to reproduce with the robot the body part of the macaw cub. In the experiment it is not important that all primitives are covered, but how mutual interaction in both directions could develop. Thus, the robot was set to start generating one of the primitives, and the human was instructed to try for each primitive by turn to cover the behavior repertory as much as possible, within a given time. The human was also instructed to proceed at will, so no predetermined order was recommended in trying to accomplish the primitives. After the experiment, the human was requested to verbally report on the primitives attempted with the robot. 

A simple proportional controller was designed to endorse the robot with the capacity of enacting deliberation and adjustment (motor compliance) in relation to the human's actions (see Fig. \ref{fig:interaction}). Thus, the position of the robot's end-effector emerging from the interaction was determined from the linear interpolation of the human and the robot desired actions, such that $position = \gamma human - (1-\gamma)robot$. For the study case $\gamma=0.9$, which means the human exerted much more influence over the position of the end-effector than the robot did. In order to avoid brusque changes in the motion trajectory, the position rate of change was saturated by a constant factor.  

In the experiment, the human interacted with VCBot during approximately 6 minutes and 20 seconds. Data was captured during 2,000 time steps, at every 100 milliseconds. The human desired actions were registered as the coordinates of the mouse cursor. In VCBot the human chooses when to enable interaction by pressing and holding the control key from the computer keyboard while moving the mouse. Therefore, both the position of the mouse cursor and the key press event were recorded. The robot desired actions were generated by the neural network. The negative ELBO  was also recorded as a measure for real-time minimization of free energy. Several repetitions of the experiment were performed, Fig. \ref{fig:signal} shows the time evolution of data captured for the most interesting trial. A total of eight human intervention events were produced during the experiment. Table \ref{tab:events} presents the human self-reported intention on the events. These results are analyzed next.

\definecolor{orangeRed}{HTML}{FF4500}
\begin{figure}[h!]
	\centering
	\begin{scriptsize}			
		\begin{tikzpicture}
		\node [] at (0,0){
			\includegraphics[width=0.47\textwidth,keepaspectratio]{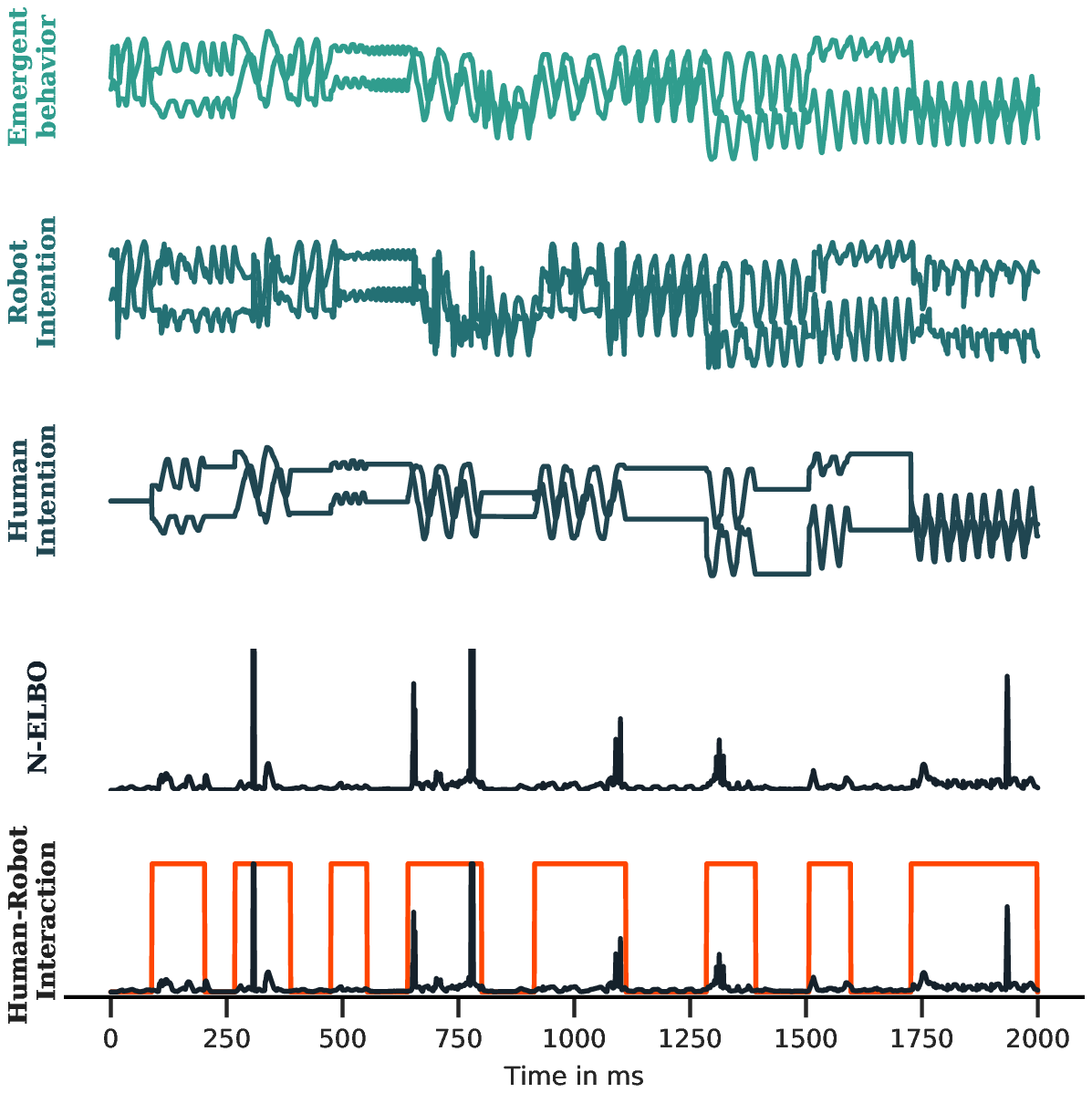}};
		\node [] at (0,-4.5){
			\includegraphics[width=0.35\textwidth,keepaspectratio]{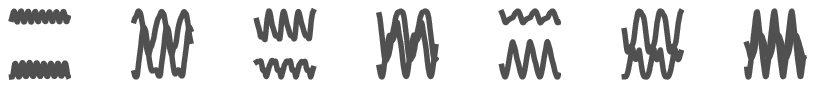}};
		
		
		\begin{scope}[shift={(0.0,0.45)}]
		\node [color=orangeRed] at (-2.65,-2.55) {\tiny 1};
		\node [color=orangeRed] at (-2.05,-2.55) {\tiny 2};
		\node [color=orangeRed] at (-1.45,-2.55) {\tiny 3};
		\node [color=orangeRed] at (-0.8,-2.55) {\tiny 4};
		\node [color=orangeRed] at (0.2,-2.55) {\tiny 5};
		\node [color=orangeRed] at (1.3,-2.55) {\tiny 6};
		\node [color=orangeRed] at (2.0,-2.55) {\tiny 7};
		\node [color=orangeRed] at (3.1,-2.55) {\tiny 8};			
		\end{scope}

		\begin{scope}[shift={(0.0,-0.5)}]
		\node [color=black] at (-2.65,-4.4) {\tiny Eye};
		\node [color=black] at (-1.7,-4.4) {\tiny Head};
		\node [color=black] at (-0.85,-4.4) {\tiny Beak};
		\node [color=black] at (0.0,-4.4) {\tiny Neck};
		\node [color=black] at (0.85,-4.4) {\tiny Right};
		\node [color=black] at (0.85,-4.6) {\tiny wing};
		\node [color=black] at (1.75,-4.4) {\tiny Belly};
		\node [color=black] at (2.65,-4.4) {\tiny Left};
		\node [color=black] at (2.65,-4.6) {\tiny wing};
		\end{scope}	
		
		\node [color=black] at (0.0,4.2) {\small \textbf{Time evolution of interaction}};
		\end{tikzpicture}
	\end{scriptsize}
	\caption{In the plots the axes and labels are not shown for clarity. The vertical component of the signals correspond for the top three plots to the width and the depth coordinates of the robot's end-effector. The N-ELBO is shown in the two lower plots. For the plot Human-Robot Intention, a scaled binary signal representing the control key pressing event by the human is also shown. Temporal contiguous human interventions were grouped and numbered. The horizontal component corresponds for all cases to the time dimension.}
	\label{fig:signal}
\end{figure}

\begin{table}[h!]
	\caption{\small{\textbf{Human intention self-report}}}
	\label{tab:events}
	\begin{center}
		\begin{tabular}{l p{7.0cm}}
			\small{\textit{E}} & \small{\textit{Description}} \\ \hline
			\small{1} & \small{The robot was doing Head, the human intended to do Beak, so the robot could switch accordingly.}\\
			\small{2} & \small{The robot was doing Beak, the human tried to do Head, which the robot accomplished.}\\
			\small{3} & \small{The robot was doing Head, the human successfully induced a change to Eye.}\\
			\small{4} & \small{The robot was doing Eye, the human tried Neck for a while, however the robot switched to Left-wing.}\\
			\small{5} & \small{The robot was doing Left-wing, the human tried again Neck, so the robot could follow up this time.}\\
			\small{6} & \small{The robot was doing Neck, the human induced a switch to Belly.}\\
			\small{7} & \small{The robot was doing Belly, the human made it change to Right-wing.}\\
			\small{8} & \small{The robot was doing Right-wing, the human tried in vain switching to Left-wing.}\\\hline
		\end{tabular}
	\end{center}
\end{table}

\subsection{Analysis}

A quantitative measure of intentionality congruence is proposed based on an automatic regression observer. The objective is classifying the robot and the human intention from the observer's evaluation, which receives as input the intended behavior signal, buffered in a limited temporal window, and outputs the attribution of the signal to a behavior category (i.e. a represented body part of the macaw cub). This problem is analogous to some extent to the comparison of time series for speech recognition (e.g. Sakoe \& Chiba \cite{Sakoe78}). Thus, two conditions are assumed in the comparison: a) patterns are time-sampled with a common and constant sampling period, and b) there is no a priori knowledge about which parts of the patterns contain important information. 

The automatic regression observer designed corresponded to a feed-forward model with the following layered structure: 10 input units (2 degrees of freedom x 5 time steps buffer window), 150 units in the first hidden layer, 100 units in the second hidden layer, 7 output units (representing the macaw cub body parts categories). The hyperbolic tangent activation function was selected for the hidden layers, and the sigmoid activation was selected for the output layer. 

For constituting the training set, the PV-RNN model generated during 200 time steps each behavior primitive from the prior distribution. Since the initial position was the same for all generations (the center of the workspace), the first 20 generation steps were discarded, in order to ensure the effector position entered the limit cycle attractor of each primitive, consequently, the training sequences had 180 time steps. 

The test set was constituted with data captured from the human. The subject was instructed to manipulate the mouse to generate in VCBot the primitives, provided a visual guide on the dataset (data was plotted in watermark color in the workspace). The confusion matrix is presented in Table \ref{tab:confusion}. As noticed, classifications were reasonably accurate for the test set.

\begin{figure}[h!]
	\centering
	\begin{scriptsize}			
		\begin{tikzpicture}
		\node [] at (0,0){
			\includegraphics[width=0.47\textwidth,keepaspectratio]{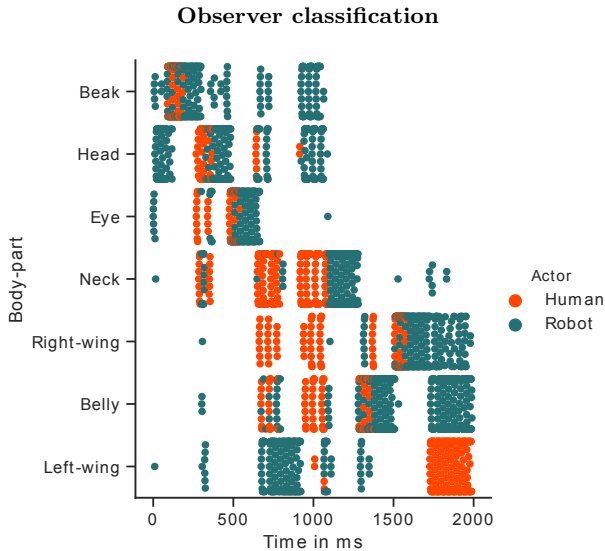}};
		\node [color=black] at (0.0,3.8) {\small \textbf{Observer classification}};
		\end{tikzpicture}
	\end{scriptsize}
	\caption{The scatter plot presents the intended action from the human and the robot actors, as evaluated by the observer. The human intervened a total of 8 times to influence the robot behavior, which is also illustrated in Fig. \ref{fig:signal} and self-reported in Table \ref{tab:events}.}
	\label{fig:observer}
\end{figure}

\begin{table}[h!]
	\caption{\small \textbf{Testing set confusion matrix}}
	\label{tab:confusion}
	\centering
	\begin{tabular}{p{0.7cm}| p{0.5cm} p{0.5cm} p{0.5cm} p{0.5cm} p{0.5cm} p{0.5cm} p{0.5cm}}
		& \scriptsize{\textit{Eye}} & \scriptsize{\textit{Head}} & \scriptsize{\textit{Beak}} & \scriptsize{\textit{Neck}} & \scriptsize{\textit{Rwing}} & \scriptsize{\textit{Belly}} & \scriptsize{\textit{Lwing}} \\\hline       
		\scriptsize{\textit{Eye}} & \scriptsize{1.000} & \scriptsize{0.000} & \scriptsize{0.000} & \scriptsize{0.000} & \scriptsize{0.000} & \scriptsize{0.000} & \scriptsize{0.000} \\          
		\scriptsize{\textit{Head}} & \scriptsize{0.000} & \scriptsize{1.000} & \scriptsize{0.000} & \scriptsize{0.000} & \scriptsize{0.000} & \scriptsize{0.000} & \scriptsize{0.000}\\          
		\scriptsize{\textit{Beak}} & \scriptsize{0.000} & \scriptsize{0.000} & \scriptsize{1.000} & \scriptsize{0.000} & \scriptsize{0.000} & \scriptsize{0.000} & \scriptsize{0.000}\\          
		\scriptsize{\textit{Neck}} & \scriptsize{0.000} & \scriptsize{0.000} & \scriptsize{0.000} & \scriptsize{0.976} & \scriptsize{0.000} & \scriptsize{0.000} & \scriptsize{0.024}\\          
		\scriptsize{\textit{Rwing}} & \scriptsize{0.000} & \scriptsize{0.000} & \scriptsize{0.000} & \scriptsize{0.000} & \scriptsize{1.000} & \scriptsize{0.000} & \scriptsize{0.000}\\          
		\scriptsize{\textit{Belly}} & \scriptsize{0.000} & \scriptsize{0.000} & \scriptsize{0.000} & \scriptsize{0.000} & \scriptsize{0.000} & \scriptsize{1.000} & \scriptsize{0.000}\\
		\scriptsize{\textit{Lwing}} & \scriptsize{0.000} & \scriptsize{0.000} & \scriptsize{0.000} & \scriptsize{0.000} & \scriptsize{0.000} & \scriptsize{0.006} & \scriptsize{0.994}\\\hline
	\end{tabular}
\end{table}

Figure \ref{fig:observer} presents the automatic regression observer's performance for the experimental data. A first aspect to be noticed is that classification for the human intention is available only during the events illustrated in Fig. \ref{fig:interaction} and self-reported in Table \ref{tab:events}, whereas data is available from the robot throughout the whole experiment (except for the first 5 time steps, given the size of the temporal buffer window). Although the interaction was subject to stochasticity from the human motions and possible incongruence between the human and the robot intentions, temporally correlated classifications for the human actions are observed around a main category and adjacent neighbors, which suggests the human likely focused on particular parts of the bird's body during the interaction events. 

In order to estimate the probability of intention congruence for an interaction episode $i$, the binomial variable $c_t \in [0,1]$ is defined from the human $\psi^\mathrm{human}$ and the robot $\psi^\mathrm{robot}$ intended actions in the time interval $[s-t^i,t^i]$, so that 

\begin{equation}
c_t=f\left(\psi^\mathrm{human}_{[s-t^i,t^i]}, \psi^\mathrm{robot}_{[s-t^i,t^i]}\right),
\label{eq:c}    
\end{equation}

\noindent where $t^i$ is the last time step of the episode, and 

\begin{equation}
f\left(\psi^\mathrm{human}_{[s-t,t]}, \psi^\mathrm{robot}_{[s-t,t]}\right) = \left\{
\begin{array}{l l}
1 &  \mathrm{if}\ O\left(\psi^\mathrm{human}_{[s-t,t]}\right) = O\left(\psi^\mathrm{robot}_{[s-t,t]}\right), \\
0 & \mathrm{otherwise}
\end{array}\right.
\label{eq:fc}    
\end{equation}

\noindent given the observer's classification function $O\left(\psi^{*}_{[s-t,t]}\right)$. 

The probability of intentional congruence for the event $i$ at time $t$ is defined such that

\begin{equation}
P\left(C^i_t|\psi^\mathrm{human},\psi^\mathrm{robot}\right) = \frac{1}{Y} \sum^{j=t}_{j=Y-t} c_t,
\label{eq:intention}    
\end{equation}

\noindent where $Y$ acts as a low pass filter to reduce classification errors due to instantaneous observation noise. 

\begin{figure}[h!]
	\centering
	\begin{scriptsize}			
		\begin{tikzpicture}
		\node [] at (0,0){
			\includegraphics[width=0.48\textwidth,keepaspectratio]{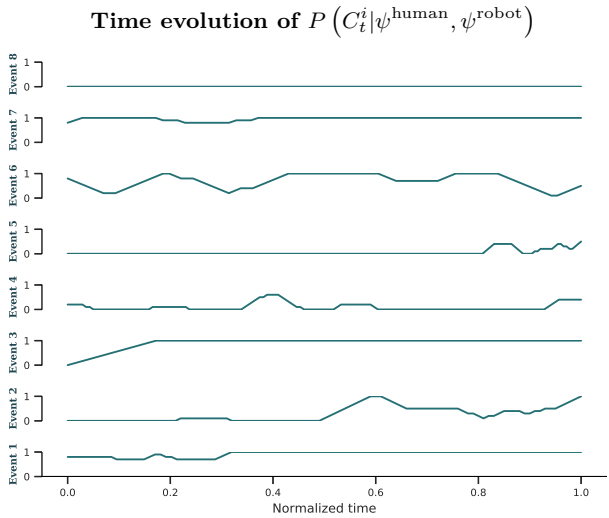}};
		\node [color=black] at (0.0,3.5) {\small \textbf{Time evolution of $P\left(C^i_t|\psi^\mathrm{human},\psi^\mathrm{robot}\right)$}};
		\end{tikzpicture}
	\end{scriptsize}
	\caption{The time dimension (horizontal axis) is normalized for the interaction events for comparisons. The vertical axis shows the probability of intentional congruence (see Eq. \ref{eq:intention}), with $Y=10$ time steps.}
	\label{fig:congProb}
\end{figure}

Figure \ref{fig:congProb} presents the time evolution of the probability of intention congruence, grouped by interaction events. As noticed, for some events congruence was observed earlier in the interaction (e.g. the events 1,3,6,7), so the human was perhaps not too sensitive to feedback from the robot and persisted longer than required to induce intention switching. On the other hand, in the $5^\mathrm{th}$ event, although the probability of intention congruence was not very high by end of the event (it was estimated to be 0.5), once the human ceased to intervene around the time step 1200 (see Fig \ref{fig:observer}), it is clear that the robot switched to the human-self reported intention (see Table \ref{tab:events}), so the human was perhaps more sensitive to feedback from the robot in this event. Finally, the events 4 and 8 were mostly characterized by intention incongruence.         

\section{Perspectives for human science}
\label{sec:perspectives}

The previous sections have presented the foundations and the proposal of a methodological resource for studying human-robot interaction, through the modeling of neural cognitive control, from artificial recurrent neural networks. An experiment was described where a human subject interacted with a virtual Cartesian robot via the computer mouse, in order to illustrate the potentialities of our approach. In this section, we argue on possible implications and perspectives for human science research. Notably, we focus on the fields of developmental psychology, education technology, and cognitive rehabilitation.

\subsection{Developmental psychology} 

The introduction section has contextualized our research interests in the domain of enactivist social cognition. Notably, in the study of \textit{primary intersubjectivity}. Thus, we have been inspired by Trevarthen's influential work where maternal-infant interactions were recorded and analyzed (Trevarthen \cite{trevarthen1979}). The work constituted a fundamental criticism to theory of a mind accounts of social cognition. It described intuitive communication in dyadic interactions as a shared control situation, where individuals reciprocally influence each other.

Although interaction theories of social cognition are not consensual, there is a shared interest in studying dyadic interaction as a second person perspective level of experience, which corresponds to forms of mutual relatedness, co-experiencing or intersubjective experiencing through reciprocal interaction. Our research is consistent with such agenda. Hence, as shown in the case study experiment, the considered scenario is characterized by direct interaction, where the human modified the body posture of the virtual robot and received visual feedback on the robot's motion. Hence, the human tried to communicate with the robot through corporal patterns or gestures. We have shown in related studies with real robots (Chame \& Tani \cite{chame2019}) that this relation can be studied both ways in the physical dimension, so the human and the robot are able to modify each other's body posture.

We have described interaction enactment as the agent's capability of taking deliberative action while conforming or adjusting to the actions induced by the partner. Consequently, behavior emerges only partly under the volitional control of the agent. We have proposed to study such dynamics from an interpretation of FEP theory which is consistent with autopoietic enactivist social cognition. Consequently, we investigate perception, cognition, and action as explained by an hermeneutic circle in dyadic encounters.

Given the context where IT emerged (i.e. opposing a cognitivist view of social cognition), assumptions based on knowledge representation and inference have been criticized when studying primary intersubjectivity. This is pointed out by Reddy \cite{reddy20184E} when analyzing assumptions about the nature and availability of mind that views early human communication as a process where the infant infers other's hidden mental states. Grounded in FEP theory, our work does not study primary intersubjectivity as a theory of mind skill. We have argued that internal representation and  prediction are considered in the  generative  sense,  from the organismic autopoietic self-organization. As pointed out by Bruineberg \& Rietveld \cite{bruineberg2019}, perception and action from anticipatory dynamics and free-energy minimization ensure the agent with the capability of maintaining adaptive sensibility  to  relevant  affordances in a given context, which would not be about reconstructing the structural hidden causes of the other's mind. Hence, the agent perceives action affordances characterized by intention, and tries to fulfill them to make sense in the interaction by taking motor action.

When analysing Fig. \ref{fig:congProb}, it is possible to notice that occasionally the human and the robot intentions were easily synchronized (e.g. in the events 1, 6, and 7 their intentions became similar early in the interaction interval). Other times, the human persisted for a while before human-to-robot alternations could take place (e.g. in the events 2, 3 and 5 the robot changed its intention according to the human's intentions). Finally, at times the human had to adapt to the robot's intentions, so robot-to-human alternations were produced (e.g. the events 4 and 8). These results present some similarities with respect to situations described in mother-infant interactions (Trevarthen \cite{trevarthen1979}). We believe that, through the design of human robot experiments of this sort, it is possible to study diverse aspects of social cognition, such as the emergence of consensus in intuitive non-verbal communication. As it was explained in the methodological description, the robot was not trained to learn how to shift from the generation of one behavior to another. Such changes resulted from the human actions on its body, and from its capacity of perceiving action affordances that make sense in the context of interaction.

Perhaps an interesting direction of research would also be studying the developmental aspect of human primary intersubjectivity skills in human-robot interaction with subjective robots. As pointed out by Torres et al. \cite{torres2013}, studying interaction is certainly subject to considerable methodological difficulties, so behavior measures of physical movements are rarely included. We believe that behavior measures could be complemented with the inclusion of the subjective dimension of the robot partner, so building more informed psychometric observations on real-time interaction skills. This can be a relevant perspective for developmental psychology, since most psychometric instruments evaluate interaction with static objects.

Through the proposal of NRL, our work with neural agents has described in detail the dynamics of action deliberation. It could be argued that our proposal could have invested some efforts in detailing the dynamics of action compliance. This aspect was left out of scope mainly due to the fact that robotic platforms are diverse. Furthermore, commonly affordable robots are designed under classical engineering modeling approach, including conventional control schemes and architectures. Since real-time responsiveness is a fundamental aspect, it is somehow problematic proposing a control model that would run over a virtual abstraction of the host native platform. Hopefully, in the near future more bio-inspired robotic structures, capable of evolving in the phylogenetic sense, would be available for most research labs. 

A final aspect to be discussed is that developmental psychology studies in mother-infant interaction have reported early behavior patterns as rudimentary, much distinct from the stylized motions proposed in the case study outlining the body shape of a macaw cub, which the artificial neural model learned. Although it is possible to model rudimentary behavior in neural agents, the inclusion of an artistic scenario is perhaps more related to research perspectives on other fields in human science, which is discussed below.

\subsection{Dynamics, consensus, and educational technology}

According to Ackermann \cite{ackermann2001}, societal convictions on the meanings of being knowledgeable or intelligent, and what it takes to become so, drive attitudes and practices in education. Hence, several theories of learning (e.g. \textit{behaviorism}, \textit{cognitivism}, \textit{constructivism}, \textit{connectivism}, \textit{constructionism}) have reflected those convictions, and have been central to the contributions of theorists in education (e.g. Bloom \cite{bloom1956}, Freire \cite{freire2018}, Siemens \cite{siemens2005}). 

In education theory, we contextualize our work within the view of \textit{constructionism}, which consists in a philosophy of learning through building artifacts in the world that reflect one's ideas. Undoubtedly, a key influential initiative in approaching technology to education took place in the late sixties, and consisted in the invention of the LOGO programming language in the Massachusetts Institute of Technology (Papert \cite{Papert1980}). The idea was to include computers in education processes, so that children learn to communicate with computers through a mathematical-logical language, in order to build their own tools and mediations to support their interest within a given context. 

\textit{Constructionism} has transcended the virtual world to ground the prototyping of physical agents through the Lego Mindstorm technology (Martin et al. \cite{martin2000}). This platform has been extensively used in the international educational robotics competition \textit{RoboCup}, notably, in the Junior league\footnote{https://www.robocup.org}, which is dedicated to young scientists mostly enrolled in secondary high-school. The Mindstorm technology has stimulated an explosion of other proposals in assemblage robotics kits.

Interesting questions are raised once new technologies are incorporated into education, in particular, inquiring the actual benefits those technologies provide to distinct academic outcomes. Thus, the aspect of how educational robotics generalizes to areas which are not closely related to the field of robotics itself has been reviewed by Benitti \cite{benitti2012}. The review found that most of the studies (around 80\%) explored physics, logic, and mathematics related topics, which may suggest a low capacity of absorption to other fields. Moreover, in the light of the increasing availability of machine learning technology (e.g. software libraries, and computational gadgets), one might question the extent to which the complexity of such technology is actually realized by the children. So in practice sophisticated automation prototypes are not simply the result of putting together a set of components which are not really understood by the student (i.e. at the modest cognitive cost of connecting black-box modules).

Our research tries to face the above criticism inspired by efforts in describing the development of cognition and action from a dynamic system theory perspective (e.g. in Thelen \& Smith \cite{thelen1994}). Hence, by focusing not only in observing how a given behavior is manifested, but on describing it as a pattern of change over time, and how such change can result from the interaction of multiple subsystems within the individual, the task, and the environment; we hope to be contributing to a reflection on phenomenon which transcends a linear logical causality understanding, to an hermeneutical description of causality that takes place with the self immersion in a feedback loop, so bringing to the foreground the notions of time, intention, stochasticity, and interaction, to a regularly perceived static world.

Perhaps the view of a non-interactive world is still deeply rooted in our society. Arts has provided us with a criticism on the static world view, though forms of expressions such as the \textit{abstract expressionist} movement. Hence, the painting technique known as \textit{action painting} (Rosenberg \cite{rosenberg1952}) reflects the physical act of painting itself, so the work is more the unfolding of an event that a picture. Here, artists employ the forces and momentum generated by their body to paint (an influential exponent in this current is Jackson Pollock, some of his works are \textit{Mural} and \textit{Lucifer}).

Returning to \textit{constructionism}, we believe the modeling of subjective robots, with which communication takes place through negotiation in shared behavior control, would enrich and extend the interaction scenario envisaged by Papert's seminal ideas. Hence, by keeping in mind the principles of \textit{roboethics} (Tzafestas \cite{tzafestas2016}), robots could be conceived as systems capable of intention. We hope this sort of synthetic agent would become relevant to other fields of knowledge, beyond educational technology, an example is rehabilitation learning which is discussed next. 

\subsection{Cognitive rehabilitation and motivation}

According to Sohlberg \& Mateer \cite{sohlberg2017}, the term \textit{cognitive rehabilitation} follows short when focusing on the aspect of remediation or compensation for decreased cognitive abilities, so the term \textit{rehabilitation of individuals with cognitive impairment} would emphasize more precisely injured individuals (i.e. acquired brain injury, and traumatic brain injury) that are and will continue to be the target of cognitive rehabilitation. In this sense, although a fundamental goal for treatment is improving and compensating cognitive abilities, a larger scope including consequences for the personal, emotional, motivational, and social dimensions of the brain injury, has been incorporated into treatment plans.   

Several evidence-based reviews have been conducted with post-stroke cognitive rehabilitation treatment for specific cognitive impairments, and concluded that although there are some evidence in support, the effectiveness of treatments has yet to be established (e.g. memory deficits in das Nair \cite{das2016}, executive dysfunction in Chung et al. \cite{chung2013}, and attention deficits in Loetscher et al. \cite{loetscher2019}). Recently, a work by Maier et al. \cite{maier2020} has proposed virtual reality as a methodology for designing a rehabilitation program in several cognitive domains conjointly, as an alternative to treating cognitive domains in isolation. Along this line, we believe an interesting perspective to follow consists in exploring rehabilitation tasks based on shared control. Our results with VCBot have illustrated the methodological possibility of such integration. 

Regarding real robots, the study of Gassert \& Dietz \cite{gassert2018}, proposes a classification for robotic rehabilitation platforms into: \textit{grounded exoskeletons}, \textit{grounded end-effector devices}, and \textit{wearable exoskeletons}. These devices are torque-controlled which allows designing diverse interaction tasks involving passive, active-assisted, and active-resisted movements, depending on the treatment goals and the patient's level of impairment. As discussed in Chame \& Tani \cite{chame2019}, when considering simultaneous deliberation and adjustment, it is also possible to profile interaction styles from the robot's capacity of taking purposeful actions, while accommodating to the human's intentions. These characteristics could enrich the availability of task repertories. In this way, neural cognitive control could become an interesting resource for rehabilitation treatments.   

A study by Gor{\v{s}}i{\v{c}} et al. \cite{gorvsivc2017} has explored the benefits of treatments based on interaction, and suggested that playing competitive games with a non-impaired partner has the potential of leading to functional improvement, when compared to conventional exercising, through an increase in motivation and exercising intensity. The study found a less pronounced effect in cooperative games but a positive effect on motivation. Although it is likely that the presence of a human partner plays a role in motivation, in the study it is reported that some subjects preferred to exercise alone. Perhaps a promising line of research would be exploring whether the fact of intersubjective interaction with a robotic partner would lead to higher life quality of subjects, notably, for participants that opted for not interacting with human partners.

Diverse additional studies could be discussed for analysing possible ways neural robotics, capable of active inference, could play a role in rehabilitation. By linking this section with the previous one, we would like to argue the relevance of \textit{constructionism} as potentially contributing to intrinsically motivated engagement of patients in planning, designing tools, and selecting goals. The treatment could also benefit from observing patient's motivation dynamics (e.g. Chame et al. \cite{chame2019dynamic}), so a conjoint planning of the treatment between the patient and the therapist is done.

\section{Conclusions}
\label{sec:conclusions}

This work started from the interest in exploring possible ways neural robots can contribute to advancing the state of the art in human science. For this, our research was contextualized within the field of enactivist social cognition, notably, in the study of control sharing in dyadic interaction, taking place in primary intersubjective communication. We proposed a methodological tool for prototyping robotics agents, modeled from free energy principle theory. Through the proposal of a demonstration program for interacting, we have shown the potentialities of our methodology for real-time human-robot interaction experiments. Finally, we discussed three main perspectives for human science. Firstly, we have argued for the inclusion of neural robotics as a resource for investigating embodied social cognition in developmental psychology. Secondly, we have discussed how our proposal is related to the theory of constructionism in education, by contributing to move from the learning of linear logical causality, to a circular understanding of causality that takes place in the subject immersion in a feedback loop when building and interacting with neural robots. Finally, we have argued on the relevance of neural robots for the field of cognitive rehabilitation, and how shared control interactive tasks could complement both treatment methodologies based on physical and virtual environments.

\section*{Acknowledgments}
\label{sec:acknowledgments}

This project was fully funded by the Okinawa Institute of Science and Technology Graduate University (OIST), and developed in the Cognitive Neurorobotics Research Unit (CNRU), which is located in 1919-1, Tancha, Onna, Kunigami District, Okinawa 904-0495. 
	
\begin{appendix}
	\section*{Appendixes}
	The following sections provide the mathematical details of PV-RNN architectures for the sake of self containment. The reader is invited to consult the work by Ahmadi \& Tani \cite{ahmadi2019} for more details. In PV-RNN two information processes are involved. The generative process follows a top-down information flow. It is in charge of anticipating the sensory state $\bm{x}_{t}$ at time $t$ from prior hidden latent representations $P_{\phi}$ within the network context $c_t$. On the other side, the inference process involves latent posterior distributions $Q_{\pi}$. It consists in a bottom-up computation flow, where the surprise signal is back propagated through time (BPTT) in the network hierarchy, within a sliding temporal window $s_t$. Algorithm \ref{algo:cogControl} describes how deliberation control is achieved in the neural robot. The minimization of free energy is equivalent to the maximization of sensory evidence lower bound (ELBO). 

\begin{algorithm}
	\caption{Deliberation control}
	\label{algo:cogControl}
	\begin{algorithmic}[0]			
		\Procedure{doInteraction}{}
		\State $t \gets 0$ 
		\State $c_{t}, \bm{x}_{t}, s_t \gets$ initialize() 
		\While {$t \leq T^\mathrm{experiment}$}
		\State $t \gets t + 1$		
		\State $\bm{x}_{t} \gets$ doGeneration($P_{\phi}$, $c_{t}$)
		\State doMotorControl($\bm{x}_{t}$)
		\If {$t > S^\mathrm{size}$}  
		\State $s_{t} \gets \left[s_{2:S^\mathrm{size}},\bm{x}_{t}\right]$
		\State $c_{t+1} \gets \mathrm{doInference}\left(Q_{\pi},s_t,c_{t-S^\mathrm{size}},S^\mathrm{size}\right)$ 				
		\Else
		\State $s_{t} \gets \left[s_{1:t},\bm{x}_{t}\right]$
		\EndIf									
		\EndWhile
		\EndProcedure
	\end{algorithmic}
	\begin{algorithmic}[0]			
		\Procedure{doInference}{$Q_\pi,s^\mathbf{y},c,n$}		
			\State $ELBO \gets \mathrm{initialize}()$
			\For {$e \gets 1, N^\mathrm{epochs}$} 					
				\State $s^\mathbf{x},s^c \gets$ doGeneration($Q_{\pi}$, $c, n$)    		
				\State $ELBO\gets$  doBPTT($s^\mathbf{x},s^\mathbf{y}, s^c$)
			\EndFor
			\State $\gets s^c_t$
		\EndProcedure
	\end{algorithmic}
\end{algorithm}

\begin{figure}[h!]
	\centering
	\begin{scriptsize}			
		\begin{tikzpicture}
		\node [] at (0,0){
			\includegraphics[width=0.4\textwidth,keepaspectratio]{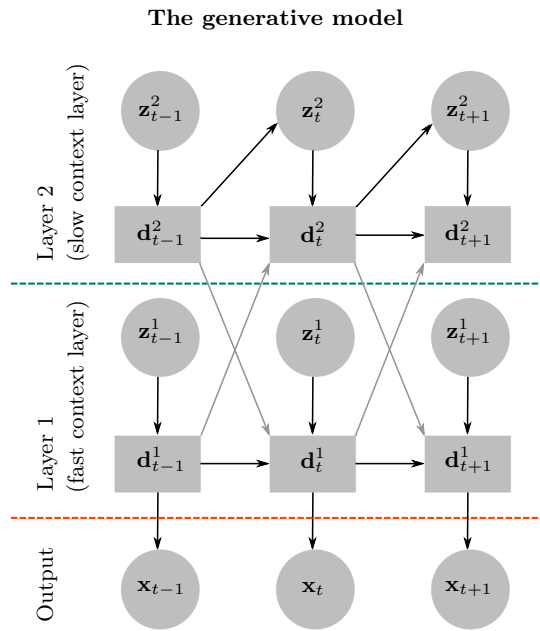}};
		\node [color=black] at (0.0,4.4) {\small \textbf{The generative model}};
		\node [rotate=90] at (-3.0,1.7) {\small Layer 2};
		\node [rotate=90] at (-2.6,2.5) {\small (slow context layer)};
		\begin{scope}[shift={(0.0,-3.0)}]
		\node [rotate=90] at (-3.0,1.65) {\small Layer 1};
		\node [rotate=90] at (-2.6,2.4) {\small (fast context layer)};
		\end{scope}
		
		
		\node [rotate=90] at (-3.0,-3.1) {\small Output};
		\node [] at (-1.5,-3.15) {\small $\mathbf{x}_{t-1}$};
		\node [] at (0.5,-3.15) {\small $\mathbf{x}_{t}$};
		\node [] at (2.55,-3.15) {\small $\mathbf{x}_{t+1}$};
		
		\begin{scope}[shift={(0.0,1.7)}]
		\node [] at (-1.5,-3.15) {\small $\mathbf{d}^{1}_{t-1}$};
		\node [] at (0.5,-3.15) {\small $\mathbf{d}^{1}_{t}$};
		\node [] at (2.55,-3.15) {\small $\mathbf{d}^{1}_{t+1}$};
		\end{scope}
		
		\begin{scope}[shift={(0.0,3.4)}]
		\node [] at (-1.5,-3.15) {\small $\mathbf{z}^{1}_{t-1}$};
		\node [] at (0.5,-3.15) {\small $\mathbf{z}^{1}_{t}$};
		\node [] at (2.55,-3.15) {\small $\mathbf{z}^{1}_{t+1}$};
		\end{scope}
		
		\begin{scope}[shift={(0.0,4.7)}]
		\node [] at (-1.5,-3.15) {\small $\mathbf{d}^{2}_{t-1}$};
		\node [] at (0.5,-3.15) {\small $\mathbf{d}^{2}_{t}$};
		\node [] at (2.55,-3.15) {\small $\mathbf{d}^{2}_{t+1}$};
		\end{scope}
		
		\begin{scope}[shift={(0.0,6.35)}]
		\node [] at (-1.5,-3.15) {\small $\mathbf{z}^{2}_{t-1}$};
		\node [] at (0.5,-3.15) {\small $\mathbf{z}^{2}_{t}$};
		\node [] at (2.55,-3.15) {\small $\mathbf{z}^{2}_{t+1}$};
		\end{scope}

		\end{tikzpicture}
	\end{scriptsize}
	\caption{Graph representation for the generative model of the PV-RNN framework (Ahmadi \& Tani \cite{ahmadi2019}) in a MTRNN setting (Yamashita \& Tani \cite{yamashita2008}). A two-level layer hierarchy is shown for illustration.}
	\label{fig:genModel}
\end{figure}

	\section*{The Generative model}
	\label{app:gen}
	
	Let\footnote{Notation: layer's latent states are denoted bold low-case, biases are denoted $\bm{b}$, weight connections are denoted $\bm{W}$ with subscripts indicating the origin and destination of the connection (e.g., $\bm{W}_\mathrm{zd}$ are the weights connecting z to d units). Superscripts $k\in\{1,...,K\}$ indicate from low to high, the layer's level in the MTRNN hierarchy. Finally, the superscripts p and q are used to distinguish between variables that belong to the prior and posterior distributions, respectively.} the generative model $P_\phi$ be defined from the parameters $\phi$, distributed among the components: generated prediction $\bm{x}$, stochastic $\bm{z}$ and the deterministic $\bm{d}$ latent states. Given the generative model of PV-RNN (see Fig. \ref{fig:genModel}), for a prediction $\bm{x}_{1:T} = (\bm{x}_{1},\bm{x}_{2},..., \bm{x}_{T})$, and considering the parameters $\phi_\mathrm{x}$, $\phi_\mathrm{z}$, and $\phi_\mathrm{d}$, $P_\phi$ factorizes such that:
	
	\begin{equation}
	\begin{multlined}
	P_\phi\left(\bm{x}_{1:T},\bm{z}_{1:T},\bm{d}_{1:T}|\bm{z}_{0},\bm{d}_{0}\right) =\\
	\prod_{t=1}^{T}P_{\phi_\mathrm{x}}\left(\bm{x}_t |\bm{d}_t\right)P_{\phi_\mathrm{z}}\left(\bm{z}_t |\bm{d}_{t-1}\right)P_{\phi_\mathrm{d}}\left(\bm{d}_t |\bm{d}_{t-1}, \bm{z}_t\right)
	\end{multlined}
	\label{eq:prior}
	\end{equation}
	
	Let the deterministic states be defined according to a MTRNN structure (Yamashita \& Tani \cite{yamashita2008}). For the $k^\mathrm{th}$ context layer at time $t$, with timescale $\iota^k$, the internal dynamics are represented such that
	
	\begin{equation}
	\label{eq:h}
	\begin{multlined}
	\bm{h}^{k}_{t} = \left(1-\frac{1}{\iota^k}\right)\bm{h}^k_{t-1}+ \frac{1}{\iota^k}\left(\bm{W}^{kk}_\mathrm{dh}\bm{d}^k_{t-1} + \bm{W}^{kk-1}_\mathrm{dh}\bm{d}^{k-1}_{t-1} \right.+ \\\left. \bm{W}^{kk+1}_\mathrm{dh}\bm{d}^{k+1}_{t-1} + \bm{W}^{kk}_\mathrm{zh}\bm{z}^k_{t} + \bm{b}^k_\mathrm{h}\right),			
	\end{multlined}
	\end{equation} 
	
	\begin{equation}
	\bm{d}^{k}_{t} = \mathrm{tanh}\left(\bm{h}^{k}_{t}\right).
	\label{eq:d}
	\end{equation} 
	
	The prior distribution $P_{\phi_\mathrm{z}}\left(\bm{z}_t|\bm{d}_{t-1}\right)$ is modeled as a Gaussian with diagonal covariance matrix, such that
	
	\begin{equation}
	P_{\phi_\mathrm{z}}\left(\bm{z}_t|\bm{d}_{t-1}\right) = \mathcal{N}\left(\bm{z}_t;\bm{\mu}^\mathrm{p}_t,\bm{\sigma}^\mathrm{p}_t\right),
	\label{eq:zp}
	\end{equation} 
	
	\noindent where $\bm{\mu}^\mathrm{p}_t$ and $\bm{\sigma}^\mathrm{p}_t$ are, respectively, the mean and standard deviation of $\bm{{z}_t} = \bm{\mu}^\mathrm{p}_{t}+\bm{\sigma}^\mathrm{p}_{t}\ast \bm{\epsilon}$, with $\bm{\epsilon}$ sampled from $\mathcal{N}(0,1)$. The variables $[\bm{\mu}^\mathrm{p}_t,\mathrm{log}\ ( \bm{\sigma}^\mathrm{p}_t)] = f_{\phi_{z}}(\bm{d}_{t-1})$ are obtained with $f_{\phi_{z}}(.)$ the one layer feed-forward neural network, such that 
	
	\begin{equation}
	\label{eq:up}
	\bm{u}^{\mathrm{p},k}_t = \bm{W}^{\mathrm{p},kk}_\mathrm{d\mu} \bm{d}^{k}_{t-1} + \bm{b}^{\mathrm{p},k}_\mathrm{\mu},
	\end{equation}
	
	\begin{equation}
	\label{eq:mup}
	\bm{\mu}^{\mathrm{p},k}_t = \mathrm{tanh}\left(\bm{u}^{\mathrm{p},k}_t\right),
	\end{equation}
	
	\begin{equation}
	\label{eq:lsp}
    \mathrm{log}\left(\bm{\sigma}^{\mathrm{p},k}_t\right) = \bm{W}^{\mathrm{p},kk}_\mathrm{d\sigma} \bm{d}^{k}_{t-1} + \bm{b}^{\mathrm{p},k}_\mathrm{\sigma}.
	\end{equation}
	
	\section*{The inference model}
	\label{app:inf}
	
	Let the inference model $Q_\pi$ (the approximate posterior) be defined from the parameters $\pi$, such that
	
	\begin{equation}
	Q_\pi(\bm{z}_t|\bm{d}_{t-1},\bm{e}_{t:T}) = \mathcal{N}\left(\bm{z}_t;\bm{\mu}^\mathrm{q}_t,\bm{\sigma}^\mathrm{q}_t\right),	
	\label{eq:posterior}
	\end{equation}
	
	\noindent where $\bm{\mu}^\mathrm{q}_t$ and $\bm{\sigma}^\mathrm{q}_t$ are, respectively, the mean and standard deviation of $\bm{{z}_t} = \bm{\mu}^\mathrm{q}_{t}+\bm{\sigma}^\mathrm{q}_{t}\ast \bm{\epsilon}$, with $\bm{\epsilon}$ sampled from $\mathcal{N}(0,1)$. The variables $[\bm{\mu}^\mathrm{q}_t,\mathrm{log}\ ( \bm{\sigma}^\mathrm{q}_t)] = f_{\pi_{z}}(\bm{d}_{t-1}, \bm{a}^\mathrm{\bar{x}})$ are obtained with $f_{\pi_{z}}(.)$ the one layer feed-forward neural network, such that 
	
	\begin{equation}
	\label{eq:uq}
	\bm{u}^{\mathrm{q},k}_t = \bm{W}^{\mathrm{q},kk}_\mathrm{d\mu} \bm{d}^{k}_{t-1} + \bm{a}^{\mathrm{\bar{x}},k}_{\mathrm{\mu},t} + \bm{b}^{\mathrm{q},k}_\mathrm{\mu},
	\end{equation}
	
	\begin{equation}
	\label{eq:muq}
	\bm{\mu}^{\mathrm{q},k}_t = \mathrm{tanh}\left(\bm{u}^{\mathrm{q},k}_t\right),
	\end{equation}
	
	\begin{equation}
	\label{eq:lsq}
	\mathrm{log}\left(\bm{\sigma}^{\mathrm{q},k}_t\right) = \bm{W}^{\mathrm{q},kk}_\mathrm{d\sigma} \bm{d}^{k}_{t-1} + \bm{a}^{\mathrm{\bar{x}},k}_{\mathrm{\sigma},t} + \bm{b}^{\mathrm{q},k}_\mathrm{\sigma}.
	\end{equation}
	
	The parameters $\bm{a}^\mathrm{\bar{x},k}_{1:T}$ are introduced to provide the network with information about the prediction error in relation to a given pattern $\bar{\bm{x}}$. Thus, $\bm{a}^\mathrm{\bar{x},k}_{1:T}$ is changed back propagating through time the prediction error $\bm{e}_{t:T}$, so information about the future steps of $\bar{\bm{x}}_{t:T}$, and existing dependencies with the current time step $t$, are captured such that
	
	\begin{equation}
	\label{eq:amu}
	\bm{a}^{\mathrm{\bar{x}},k}_{\mathrm{\mu},t} = \bm{a}^{\mathrm{\bar{x}},k}_{\mathrm{\mu},t} + \alpha\frac{\partial L}{\partial \bm{a}^{\mathrm{\bar{x}},k}_{\mathrm{\mu},t}},
	\end{equation}

	\begin{equation}
	\label{eq:als}
	\bm{a}^{\mathrm{\bar{x}},k}_{\mathrm{\sigma},t} = \bm{a}^{\mathrm{\bar{x}},k}_{\mathrm{\sigma},t} + \alpha\frac{\partial L}{\partial \bm{a}^{\mathrm{\bar{x}},k}_{\mathrm{\sigma},t}},
	\end{equation}
	
	\noindent with $\alpha$ denoting the learning rate. 
	
	Let the Variational Evidence Lower Bound (ELBO) $L(\phi,\pi)$ be defined by
	
	\begin{equation}
	\begin{multlined}
	L(\phi,\pi) = \sum_{t=1}^{T}\left( \frac{1}{n_\mathrm{x}}E_{\mathrm{q_\pi}}\left[\mathrm{log}\left(\bm{x}_t|\tilde{\bm{d}}_{t},\bm{z}_t\right)\right] -\right. \\
	\left.\frac{w}{n_\mathrm{z}}\mathrm{KL}\left[Q_\mathrm{\pi}\left(\bm{z}_{t}|\tilde{\bm{d}}_{t-1},\bm{e}_{t:T}\right)\|P_{\phi_z}\left(\bm{z}_t|\tilde{\bm{d}}_{t-1}\right)\right]\right).
	\end{multlined}
	\label{eq:elbo}
	\end{equation}
	
	\noindent Here, $n_\mathrm{x}$ is the number of degrees of freedom of the robot, and $n_\mathrm{z}$ is the total number of z units considering all layers. Since $\bm{d}_t$ is deterministic given $\bm{d}_{t-1}$ and $\bm{z}_t$, $\tilde{\bm{d}}_t$ denotes the center of a Dirac distribution. The first term at the right of the equation is a reconstruction component, it corresponds to the expected log-likelihood under the posterior distribution $Q_\mathrm{\pi}$. The second therm is a regulation component, it corresponds to the Kullback-Leibler (KL) divergence between the prior and the posterior distributions of the latent variables. The meta-parameter $w$ adjusts the optimization weight in learning the posterior and the prior distributions. Given that the prior and posterior distributions are Gaussian and after dropping the random variable notation to improve readability, the KL component can be expressed as

	\begin{equation}
	\begin{multlined}
	\mathrm{KL}\left[Q_\mathrm{\pi}\|P_{\phi_z}\right] =
	\mathrm{log}\left(\frac{\bm{\sigma}^\mathrm{p}}{\bm{\sigma}^\mathrm{q}}\right) + \frac{\left(\bm{\mu}^\mathrm{p}-\bm{\mu}^\mathrm{q}\right)^2 + \left(\bm{\sigma}^\mathrm{q}\right)^2}{2\left(\bm{\sigma}^\mathrm{p}\right)^2}-\frac{1}{2}
	\end{multlined}.
	\label{eq:klDiv}
	\end{equation}
	
	Finally, the $i^{\mathrm{th}}$ output dimension of the cognitive control space is obtained from the one layer feed-forward propagation of the $\bm{d}_{i,t}$ units, such that
	
	\begin{equation}
	\bm{o}_{i,t} = \bm{W}_{\mathrm{dx}_i}\bm{d}^{1}_{t} + \bm{b}_{\mathrm{x}_i}.
	\label{eq:o}
	\end{equation}
	
	\noindent Unlike in Ahmadi \& Tani \cite{ahmadi2019}, in this work it is not included in $\bm{o}_{i,t}$ connections from the stochastic latent distributions at the Low level, in order to reduce the computational complexity. Hence, the output $\bm{x}_{i,t}$ is such that
	\begin{equation}
	\bm{x}_{i,t} = \mathrm{softmax}\left(\bm{o}_{i,t}\right).
	\label{eq:softmax}
	\end{equation}
	
	Free energy is optimized for distinct purposes during the phases \textit{training} and \textit{experiment} (see Fig. \ref{fig:methodology}). The agent is trained for behavior acquisition in a supervised manner on the behavior primitives. Data can be obtained analytically from mathematically modeling behavior, or captured from direct interaction (e.g. by kinesthetic teaching). Variational models are better at generalizing when compared to deterministic frameworks, so it is not necessary to collect numerous samples on the desired behavior. During the \textit{modeling} phase, the model's meta-parameters are selected to modulate stochasticity (see Table \ref{tab:paramsModel}), resulting in a higher capability of generalization. Free energy optimization in the training phase involves the off-line modification of parameters including synaptic weights, biases, and state variables. Contrarily, the \textit{experiment} phase is characterized by on-line inference, where only state variables within the temporal sliding window are optimized (see Fig. \ref{fig:interaction}). The objective of free-energy optimization during experiment is to allow the agent to perceive action affordances in intentional interaction, through changes in the overall PV-RNN latent state. Next, the model gradients for the BPTT algorithm are provided. 
	
	\section*{BPTT gradients}
	\label{app:BPTT}
	
	In this section the back propagation trough time gradients are provided. 
	The variable change $\bm{\rho} = \mathrm{log}(\bm{\sigma})$ is proposed to improve readability. By applying the derivation chain rule, the gradients are computed from Eq. \eqref{eq:elbo}, such as:
	
	\begin{equation}
	\frac{\partial\bm{L}_{i,t}}{\partial\bm{o}_{i,t}} = \frac{1}{n_\mathrm{x}}\left(\bm{x}_{i,t}-\bar{\bm{x}}_{i,t}\right),
	\label{eq:bptt_L}
	\end{equation}
	
	\begin{equation}
	\label{eq:bptt_o}
	\begin{split}
	\frac{\partial\bm{L}_{t}}{\partial\bm{d}^{k}_{i,t-1}} = 
	\sum_{s}^{} \frac{\partial\bm{L}_{t}}{\partial\bm{o}_{s,t-1}}
	\frac{\partial\bm{o}_{s,t-1}}{\partial\bm{d}^{k}_{i,t-1}} +
	\sum_{j}^{}
	\frac{\partial\bm{L}_{t}}{\partial\bm{h}^{k}_{j,t}}
	\frac{\partial\bm{h}^{k}_{j,t}}{\partial\bm{d}^{k}_{i,t-1}}+\\
	\sum_{y}^{} \frac{\partial\bm{L}_{t}}{\partial\bm{h}^{k-1}_{y,t}}
	\frac{\partial\bm{h}^{k-1}_{y,t}}{\partial\bm{d}^{k}_{i,t-1}} +
	\sum_{r}^{} \frac{\partial\bm{L}_{i,t}}{\partial\bm{h}^{k+1}_{r,t}}
	\frac{\partial\bm{h}^{k+1}_{r,t}}{\partial\bm{d}^{k}_{i,t-1}} +\\
	\sum_{m}^{}
	\frac{\partial\bm{L}_{t}}{\partial\bm{\mu}^{\mathrm{p},k}_{m,t}}
	\frac{\partial\bm{\mu}^{\mathrm{p},k}_{m,t}}{\partial\bm{d}^{k}_{i,t-1}}
	+\sum_{m}^{}\frac{\partial\bm{L}_{t}}{\partial\bm{\rho}^{\mathrm{p},k}_{m,t}}
	\frac{\partial\bm{\rho}^{\mathrm{p},k}_{m,t}}{\partial\bm{d}^{k}_{i,t-1}}+
	\\\sum_{m}^{}\frac{\partial\bm{L}_{t}}{\partial\bm{\mu}^{\mathrm{q},k}_{m,t}}
	\frac{\partial\bm{\mu}^{\mathrm{q},k}_{m,t}}{\partial\bm{d}^{k}_{i,t-1}}
	+\sum_{m}^{}\frac{\partial\bm{L}_{t}}{\partial\bm{\rho}^{\mathrm{q},k}_{m,t}}
	\frac{\partial\bm{\rho}^{\mathrm{q},k}_{m,t}}{\partial\bm{d}^{k}_{i,t-1}}
	\end{split}.
	\end{equation}
	
	The terms involving other levels of the hierarchy are set to zero when not defined (i.e. in the case of level $k+1$ in the highest layer or the level $k-1$ in the lowest layer). Likewise, the gradient below is only defined for $k=1$
		
	\begin{equation}
	\label{eq:bptt_ot1_dt1}
	\frac{\partial\bm{o}_{s,t-1}}{\partial\bm{d}^{k}_{i,t-1}} = \bm{W}^{kk}_{\mathrm{dx}_{s,i}},
	\end{equation}
	
	\begin{equation}
	\label{eq:bptt_Lt_ht1}
	\frac{\partial\bm{L}_{t}}{\partial\bm{h}^{k}_{i,t-1}}=
	\frac{\partial\bm{L}_{t}}{\partial\bm{d}^{k}_{i,t-1}}
	\frac{\partial\bm{d}^{k}_{i,t-1}}{\partial\bm{h}^{k}_{i,t-1}}+
	\frac{\partial\bm{L}_{t}}{\partial\bm{h}^{k}_{i,t}}
	\frac{\partial\bm{h}^{k}_{i,t}}{\partial\bm{h}^{k}_{i,t-1}},
	\end{equation}

    \begin{equation}
	\label{eq:bptt_dt1_ht1}
	\frac{\partial\bm{d}^{k}_{i,t-1}}{\partial\bm{h}^{k}_{i,t-1}} =
	1- \mathrm{tanh}^2(\bm{h}^{k}_{i,t}),
	\end{equation}

	\begin{equation}
	\label{eq:bptt_ht_ht1}
	\frac{\partial\bm{h}^{k}_{i,t}}{\partial\bm{h}^{k}_{i,t-1}} =
	1- \frac{1}{\iota^{k}},
	\end{equation}
	
	\begin{equation}
	\label{eq:bptt_ht_dt1}
	\frac{\partial\bm{h}^{k}_{j,t}}{\partial\bm{d}^{k}_{i,t-1}} =
	\dfrac{1}{\iota^{k}}
	\bm{W}^{kk}_{\mathrm{hd}_{j,i}},
	\end{equation}
	
	\begin{equation}
	\label{eq:bptt_dkm1t_dt1}
	\frac{\partial\bm{h}^{k-1}_{y,t}}{\partial\bm{d}^{k}_{i,t-1}} = 
	\dfrac{1}{\iota^{k-1}}
	\bm{W}^{kk-1}_{\mathrm{dh}_{y,i}},
	\end{equation}

	\begin{equation}
	\label{eq:bptt_dkp1t_dt1}
	\frac{\partial\bm{h}^{k+1}_{r,t}}{\partial\bm{d}^{k}_{i,t-1}} = 
	\dfrac{1}{\iota^{k+1}}
	\bm{W}^{kk+1}_{\mathrm{dh}_{r,i}},
	\end{equation}
	
    \begin{equation}
	\label{eq:bptt_Lt_mupt}
	\frac{\partial\bm{L}_{t}}{\partial\bm{\mu}^{\mathrm{p},k}_{m,t}}=
	\frac{w^{k}}{n_\mathrm{z}}\left(
	\frac{\bm{\mu}^{\mathrm{p},k}_{m,t}-\bm{\mu}^{\mathrm{q},k}_{m,t}}{(\bm{\sigma}^{\mathrm{p},k}_{m,t})^2}\right),
	\end{equation}

	\begin{equation}
	\label{eq:bptt_mupt_dt1}
	\frac{\partial\bm{\mu}^{^\mathrm{p},k}_{m,t}}{\partial\bm{d}^{k}_{i,t-1}}=
	\left(1- \mathrm{tanh}^2\left(\bm{u}^{\mathrm{p},k}_{m,t}\right)\right)\bm{W}^{kk}_{{\mu^{\mathrm{p}}d}_{m,i}},
	\end{equation}
	
	\begin{equation}
	\label{eq:bptt_Lt_rhopt}
	\frac{\partial\bm{L}_{t}}{\partial\bm{\rho}^{\mathrm{p},k}_{m,t}}=
	\frac{w^{k}}{n_\mathrm{z}}
	\left(1- \frac{	\left(
	\left(\bm{\mu}^{\mathrm{q},k}_{m,t}-\bm{\mu}^{\mathrm{p},k}_{m,t}\right)^2
	+ \left(\bm{\sigma}^{\mathrm{q},k}_{m,t}\right)^2\right)}{\left(\bm{\sigma}^{\mathrm{p},k}_{m,t}\right)^2}\right),
	\end{equation}

	\begin{equation}
	\label{eq:bptt_rhoupt_dt1}
	\frac{\partial\bm{\rho}^{\mathrm{p},k}_{m,t}}{\partial\bm{d}^{k}_{i,t-1}}=
	\bm{W}^{kk}_{\mathrm{\sigma^p d}_{m,i}},
	\end{equation}
	
	\begin{equation}
	\label{eq:bptt_Lt_muqt}
	\frac{\partial\bm{L}_{t}}{\partial\bm{\mu}^{\mathrm{q},k}_{m,t}}=
	\frac{\partial\bm{L}_{t}}{\partial\bm{z}^{\mathrm{q},k}_{m,t}}+
	\frac{1}{n_\mathrm{z}}\frac{\bm{\mu}^{\mathrm{q},k}_{m,t}-\bm{\mu}^{\mathrm{p},k}_{i,t}}{\left(\sigma^{\mathrm{p},k}_{m,t}\right)^2},
	\end{equation}

    \begin{equation}
	\label{eq:bptt_Lt_zqt}
	\frac{\partial\bm{L}_{t}}{\partial\bm{z}^{\mathrm{q},k}_{m,t}}=
	\sum_{j}^{}\frac{\partial\bm{L}_{t}}{\partial\bm{h}^{k}_{m,t}}
	\left(\frac{1}{\iota^{k}}
	\bm{W}^{kk}_{\mathrm{zh}_{m,i}}\right),
	\end{equation}

    \begin{equation}
	\label{eq:bptt_muqt_dt1}
	\frac{\partial\bm{\mu}^{^\mathrm{q},k}_{m,t}}{\partial\bm{d}^{k}_{i,t-1}}=
	\left(1- \mathrm{tanh}^2\left(\bm{u}^{\mathrm{q},k}_{m,t}\right)\right)\bm{W}^{kk}_{{\mu^{\mathrm{q}}d}_{m,i}},
	\end{equation}
	
	\begin{equation}
	\label{eq:bptt_Lt_rhoqt}
	\frac{\partial\bm{L}_{t}}{\partial\bm{\rho}^{\mathrm{q},k}_{m,t}}=
	\frac{\partial\bm{L}_{t}}{\partial\bm{z}^{\mathrm{q},k}_{m,t}}
	\bm{\epsilon}^{\mathrm{q},k}_{m,t}\bm{\sigma}^{\mathrm{q},k}_{m,t}+
	\frac{w^{k}}{n_\mathrm{z}}\left(-1 +\frac{ (\bm{\sigma}^{\mathrm{q},k}_{m,t})^2}{(\bm{\sigma}^{\mathrm{p},k}_{m,t})^2}\right),
	\end{equation}
	
	\begin{equation}
	\label{eq:bptt_rhouqt_dt1}
	\frac{\partial\bm{\rho}^{\mathrm{q},k}_{m,t}}{\partial\bm{d}^{k}_{i,t-1}}=
	\bm{W}^{kk}_{\mathrm{\sigma^q d}_{m,i}},
	\end{equation}
		
	\begin{equation}
	\label{eq:bptt_auq_kk_t}
	\frac{\partial\bm{L}_{t}}{\bm{a}^{\mathrm{\bar{x}},k}_{\mathrm{\mu}_{i,t}}} = \left(1- \mathrm{tanh}^2\left(\bm{u}^{\mathrm{q},k}_{i,t}\right)\right)\frac{\partial\bm{L}_{t}}{\partial\bm{\mu}^{\mathrm{q},k}_{i,t}},
	\end{equation}

	\begin{equation}
	\label{eq:bptt_alq_kk_t}
	\frac{\partial\bm{L}_{t}}{\bm{a}^{\mathrm{\bar{x}},k}_{\mathrm{\sigma}_{i,t}}} = \frac{\partial\bm{L}_{t}}{\partial\bm{\rho}^{\mathrm{q},k}_{i,t}}.
	\end{equation}

	The synaptic weights and biases are updated such that
	
	\begin{equation}
	\label{eq:bptt_Wdh_kk_t}
	\frac{\partial\bm{L}_{t}}{\partial\bm{W}^{kk}_{\mathrm{dh}_{i,j}}}= \sum_{t} \frac{1}{\iota^{k}}\bm{d}^{k}_{j,t-1}\frac{\partial\bm{L}_{t}}{\partial\bm{h}^{k}_{i,t}}
	\end{equation}
	
    \begin{equation}
	\label{eq:bptt_bh_kk_t}
	\frac{\partial\bm{L}_{t}}{\partial\bm{b}^{k}_{\mathrm{h}_{i}}}= \sum_{t} \frac{1}{\iota^{k}}\frac{\partial\bm{L}_{t}}{\partial\bm{h}^{k}_{i,t}}
	\end{equation}

	\begin{equation}
	\label{eq:bptt_Wdh_kkm1_t}
	\frac{\partial\bm{L}_{t}}{\partial\bm{W}^{kk-1}_{\mathrm{dh}_{i,j}}}= \sum_{t} \frac{1}{\iota^{k}}\bm{d}^{k-1}_{j,t-1}\frac{\partial\bm{L}_{t}}{\partial\bm{h}^{k}_{i,t}}
	\end{equation}
	
	\begin{equation}
	\label{eq:bptt_Wdh_kkp1_t}
	\frac{\partial\bm{L}_{t}}{\partial\bm{W}^{kk+1}_{\mathrm{dh}_{i,j}}}= \sum_{t} \frac{1}{\iota^{k}}\bm{d}^{k+1}_{j,t-1}\frac{\partial\bm{L}_{t}}{\partial\bm{h}^{k}_{i,t}}
	\end{equation}
	
    \begin{equation}
	\label{eq:bptt_Wzh_kk_t}
	\frac{\partial\bm{L}_{t}}{\partial\bm{W}^{\mathrm{p},kk}_{\mathrm{zh}_{i,j}}}= \sum_{t} \frac{1}{\iota^{k}}\bm{z}^{k}_{j,t}\frac{\partial\bm{L}_{t}}{\partial\bm{h}^{k}_{i,t}}
	\end{equation}
	
    \begin{equation}
	\label{eq:bptt_Wdup_kk_t}
	\frac{\partial\bm{L}_{t}}{\partial\bm{W}^{kk}_{\mathrm{d\mu^\mathrm{p}}_{i,j}}}= \sum_{t} \left(1- \mathrm{tanh}^2\left(\bm{u}^{\mathrm{p},k}_{i,t}\right)\right)\bm{d}^{k}_{j,t-1}\frac{\partial\bm{L}_{t}}{\partial\bm{\mu}^{\mathrm{p},k}_{i,t}}
	\end{equation}
	
	\begin{equation}
	\label{eq:bptt_bup_kk_t}
	\frac{\partial\bm{L}_{t}}{\partial\bm{b}^{\mathrm{p},k}_{\mathrm{\mu}_{i}}}= 
	\sum_{t} \left(1- \mathrm{tanh}^2\left(\bm{u}^{\mathrm{p},k}_{i,t}\right)\right)\frac{\partial\bm{L}_{t}}{\partial\bm{\mu}^{\mathrm{p},k}_{i,t}},
	\end{equation}
	
	\begin{equation}
	\label{eq:bptt_Wdlp_kk_t}
	\frac{\partial\bm{L}_{t}}{\partial\bm{W}^{\mathrm{p},kk}_{\mathrm{d\sigma^\mathrm{p}}_{i,j}}}= \sum_{t} \bm{d}^{k}_{j,t-1}\frac{\partial\bm{L}_{t}}{\partial\bm{\rho}^{\mathrm{p},k}_{i,t}},
	\end{equation}

    \begin{equation}
	\label{eq:bptt_blp_kk_t}
	\frac{\partial\bm{L}_{t}}{\partial\bm{b}^{\mathrm{p},k}_{\mathrm{\sigma}_{i}}}= 
	\sum_{t} \frac{\partial\bm{L}_{t}}{\partial\bm{\rho}^{\mathrm{p},k}_{i,t}}.
	\end{equation}

The gradients $\frac{\partial\bm{L}_{t}}{\partial\bm{W}^{\mathrm{p},kk}_{\mathrm{d\mu}_{i,j}}}$, $\frac{\partial\bm{L}_{t}}{\partial\bm{b}^{\mathrm{q},k}_{\mathrm{\mu}_{i}}}$, $\frac{\partial\bm{L}_{t}}{\partial\bm{W}^{\mathrm{q},kk}_{\mathrm{d\sigma}_{i,j}}}$, and $\frac{\partial\bm{L}_{t}}{\partial\bm{b}^{\mathrm{q},k}_{\mathrm{\sigma}_{i}}}$ are computed, analogous to Eqs. \eqref{eq:bptt_Wdup_kk_t}, \eqref{eq:bptt_bup_kk_t}, \eqref{eq:bptt_Wdlp_kk_t}, and \eqref{eq:bptt_blp_kk_t}. 

\end{appendix}

	
\bibliographystyle{ieeetr}
{\small
	\bibliography{references}}

	\end{document}